\documentclass[lettersize,journal]{IEEEtran}
\usepackage{amsmath,amsfonts}
\usepackage{algorithmic}
\usepackage{algorithm}
\usepackage{array}
\usepackage[caption=false,font=normalsize,labelfont=sf,textfont=sf]{subfig}
\usepackage{textcomp}
\usepackage{stfloats}
\usepackage{url}
\usepackage{verbatim}
\usepackage{graphicx}
\usepackage{cite}
\usepackage{xcolor}
\usepackage{tabularx}

\usepackage{amsmath}
\usepackage{amssymb}
\usepackage{multirow}
\usepackage{booktabs}
\usepackage{colortbl}
\usepackage{pifont}
\graphicspath{{./Image/}}

\begin{document}

\title{SeeRF: Learning to See Beyond Fields of View via Generative Image Registration and Fusion}

\author{Xunpeng Yi, Yibing Zhang, Qinglong Yan, Yifan Lu, Han Xu and Jiayi Ma

\thanks{This work was supported by the National Natural Science Foundation of China under Grant no. 625B2135 and 62676289. \emph{(Corresponding author: Jiayi Ma.)}}% <-this % stops a space
\thanks{Xunpeng Yi, Yibing Zhang, Qinglong Yan, Yifan Lu are with the Electronic Information School, Wuhan University, Wuhan 430072, China (e-mail: yixunpeng@whu.edu.cn, zhangyibing@whu.edu.cn, qinglong\_yan@whu.edu.cn, yifanlu2024@gmail.com).}
\thanks{Han Xu is with the School of Automation, Southeast University, Nanjing 210096, China (e-mail: xu\_han@seu.edu.cn).}
\thanks{Jiayi Ma is with the Electronic Information School and the School of Robotics, Wuhan University, Wuhan, 430072 (e-mail: jyma2010@gmail.com).}
}

% The paper headers
\markboth{IEEE Transactions}%
%\markboth{IEEE Transactions on Latex}
{Yi \MakeLowercase{\textit{et al.}}: Title}

\maketitle

\begin{abstract}
Due to the inherent discrepancies in imaging viewpoints and field-of-view mismatches among multi-modal sensors, existing fusion methods have long struggled to achieve complete scene-level information integration beyond the limitation in sensor fields of view. Consequently, the generated fusion results often contain modality-specific regions and fusion artifacts, limiting the comprehensive representation of scene structures and semantic information. To address this challenge, we propose a generative registration–fusion framework, termed SeeRF. Specifically, it first introduces a deformation generation strategy based on the diffeomorphic stationary velocity field, combined with visual foundation model-guided feature extraction and a coarse-to-fine progressive multi-modal image registration module, to achieve accurate cross-modal flow estimation and generate valid field-of-view masks. Subsequently, the registration module is tightly coupled with the generative diffusion module for multi-modal completion and fusion, where the cross-modal control encoding module and the control residual feature encoder–decoder mechanism provide effective guidance. By leveraging the powerful prior modeling capability of generative diffusion models, the proposed framework overcomes the limitations imposed by shared fields of view and enables effective completion and natural fusion of non-overlapping field-of-view regions. Extensive experiments on multiple public datasets and real-world scenarios demonstrate that the proposed method achieves superior performance, validating its effectiveness, superiority, and robustness. The code will be available at https://github.com/XunpengYi/SeeRF.
\end{abstract}


\section{Introduction}
\IEEEPARstart{I}{mage} fusion is one of the fundamental problems in digital image processing~\cite{ma2020ddcgan, sun2022detfusion, liu2024task, bai2025refusion}. Single-modal images are limited by their observation dimensions, resulting in insufficient information representation. In contrast, multi-modal images can integrate complementary information from different modalities through fusion to produce a more comprehensive representation of the scene. Therefore, it has wide applications in fields such as surveillance and security, autonomous driving, and industrial inspection~\cite{liu2022target, he2023degradation, wu2025every, yi2025artificial}.

\begin{figure}[!t]
\centering
\includegraphics[width=1\linewidth]{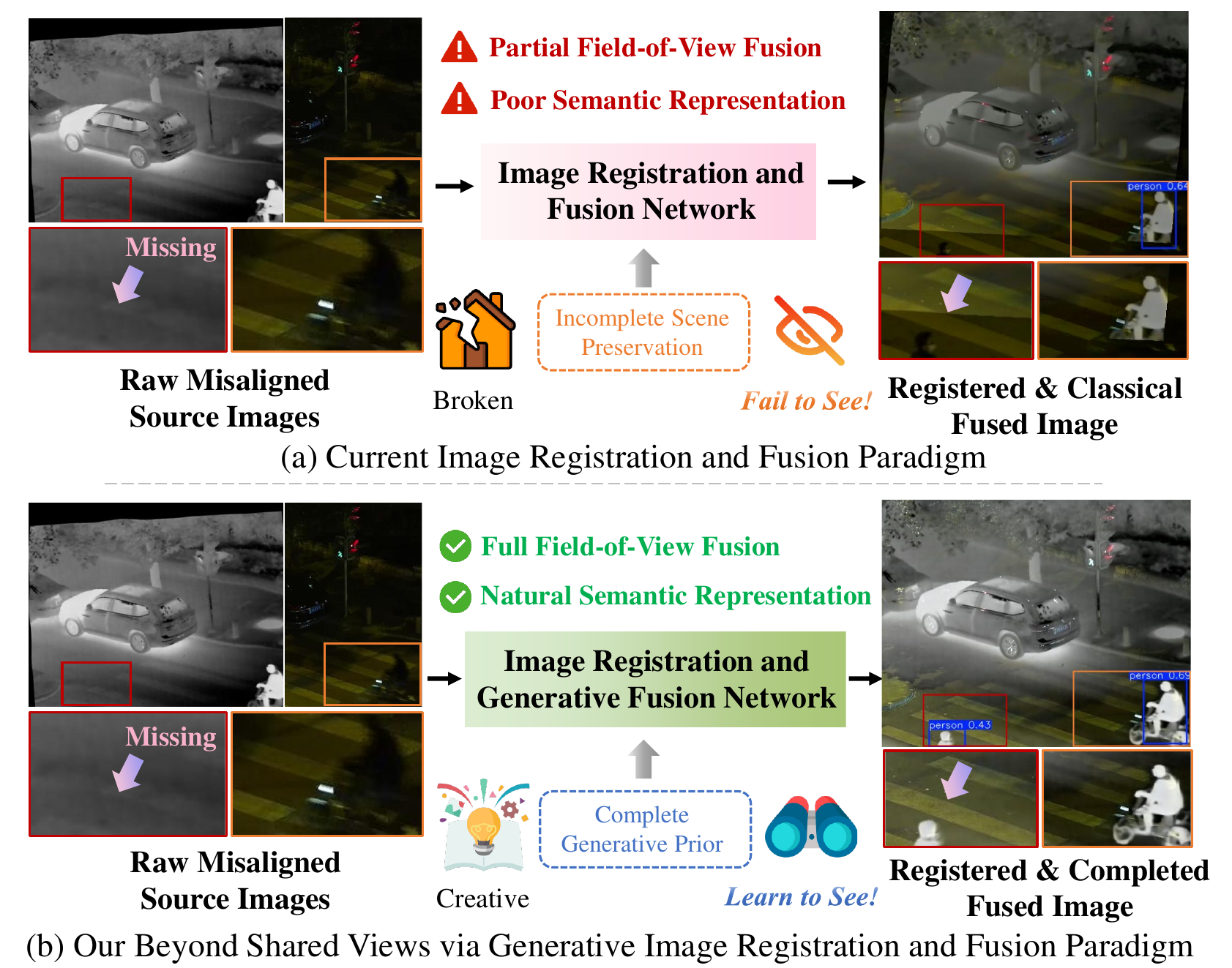}
\caption{Analysis of image registration and fusion paradigms in terms of viewpoint and field of view.}
\label{fig1-1}
\end{figure}

Infrared and visible image fusion represents a significant subfield of multi-modal image fusion. It seeks to integrate the rich textural details present in visible images with the thermal radiation characteristics captured by infrared imagery, thereby generating a unified and comprehensive scene representation. Nevertheless, the realization of such fusion fundamentally depends on the acquisition of data from both infrared and visible imaging sensors. Owing to inherent differences in sensor configurations and imaging viewpoints, existing methods characterize the problem by noting that source images acquired by multimodal sensors often exhibit substantial parallax~\cite{xu2022rfnet, xu2023murf}. Therefore, several studies have attempted to jointly address image registration and fusion~\cite{lu2025net, li2025mulfs, xu2026iterrf} to mitigate these challenges. Despite these advances, most existing joint registration–fusion methods primarily focus on correcting spatial misalignment between cross-modal images while assuming that the source images have consistent field-of-view coverage. However, this assumption does not hold under most conditions. Consequently, existing methods must tolerate such field-of-view inconsistencies and often produce degraded fusion results with noticeable artifacts, while failing to preserve complementary cross-modal information across the entire field of view, as illustrated in Fig.~\ref{fig1-1}. This also compromises the semantic integrity of the fused representation, ultimately hindering the performance of downstream tasks.

\begin{figure*}[!t]
\centering
\includegraphics[width=1\linewidth]{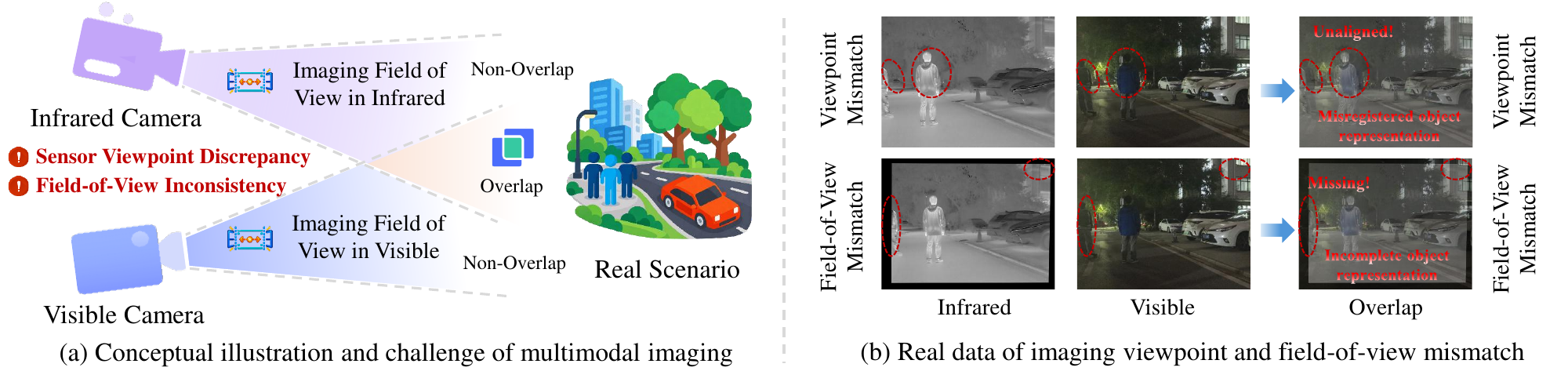}
\caption{Schematic illustration of the challenges and analysis of real data in the existing image registration and fusion field. (a) Conceptual illustration and challenge of multimodal imaging. (b) Real data of imaging viewpoint and field-of-view mismatch.}
\label{fig1-2}
\end{figure*}

On the basis of the above observations, considering differences in sensor installation positions, imaging geometry, and device intrinsic parameters, we rethink the challenges associated with data acquired by heterogeneous imaging sensors into two categories: viewpoint misalignment and field-of-view inconsistency. As illustrated in Fig.~\ref{fig1-2}, viewpoint misalignment introduces spatial displacements that disrupt spatial correspondences between cross-modal images, resulting in edge misalignment and ghosting artifacts. Meanwhile, the issues caused by field-of-view inconsistency become apparent as spatial misalignment is corrected and cross-modal alignment is established. Specifically, field-of-view inconsistency leads to non-overlapping regions and the absence of information from certain modalities, making it difficult for conventional fusion strategies to reliably integrate information in regions not covered by both sensors.

In this context, existing methods either fail to integrate information from the missing modality in regions outside the shared field of view, resulting in discontinuities between regions and inconsistencies in the distribution of fused information, or directly reduce the image coverage through center cropping, discarding field-of-view regions with degraded fusion quality and thereby producing fusion results with limited spatial coverage. This unresolved issue continues to hinder the full exploitation of the complementary information available in unregistered multimodal images for a long time.

To this end, we propose a unified framework for generative image registration and fusion that learns to see information beyond the shared field of view, termed SeeRF. In detail, we first develop a stationary-velocity-field driven multi-modal image registration module (MIR). It synthesizes an invertible deformation field and jointly incorporates a fine-tuned feature extractor and matcher, enabling progressive estimation of the flow field and achieving view alignment. Subsequently, we develop a generative diffusion module for multi-modal completion and fusion (MCF). By performing cross-modal guided completion and fusion, the model establishes consistent observations in missing-scene regions and thereby achieves field-of-view alignment. Importantly, MIR and MCF are intrinsically coupled, jointly breaking the limitations of image fusion beyond the fields of view. MIR establishes a geometrically consistent coordinate system and provides registration-confidence masks, thereby laying the foundation for MCF-guided generation. Conversely, MCF exploits its powerful generative and completion capabilities to compensate for the missing correspondences encountered by MIR in absent or non-overlapping regions. Extensive experimental results demonstrate that the proposed method effectively addresses image fusion beyond the field of view perspective, achieving superior performance and robustness across multiple datasets.

Overall, our contributions can be summarized as follows:
\begin{itemize}
  \item[-] From the perspective of multimodal sensor imaging, we propose a unified image fusion framework that jointly addresses spatial misalignment and field-of-view inconsistency. To the best of our knowledge, this is the first image fusion framework to explicitly explore and address the challenges of field-of-view discrepancies.
  \item[-] We propose a diffeomorphic stationary velocity field deformation generation strategy, which introduces the feature representation capability guided by foundation models and combines it with a coarse-to-fine progressive flow field regression mechanism to achieve high-precision cross-modal deformation modeling and flow field estimation.
  \item[-] By fully exploiting the powerful capability of generative diffusion models and combining it with the valid field-of-view mask and the collaborative effect of the registration module, we achieve effective generative completion and fusion of missing information outside the field of view, enabling the model to fully explore and utilize complementary information beyond the shared field of view across sensors.
  \item[-] Extensive experimental results demonstrate that SeeRF achieves excellent performance in both cross-modal image registration and fusion tasks, and effectively improves the performance of downstream semantic tasks.
\end{itemize}

\section{Related Work}
\subsection{Classical Image Fusion Methods}
Early image fusion methods typically relied on hand-crafted fusion strategies, resulting in inherent limitations in both fusion effectiveness and generalization capability. With the development of deep learning, image fusion has achieved substantial advances in both methodological frameworks and technical capabilities. Initially, the autoencoder-based image fusion methods were proposed to achieve feature-level fusion by using encoders and decoders pretrained on large-scale datasets, together with hand-crafted fusion rules~\cite{liu2016image, li2018densefuse}. To overcome the limitations of hand-crafted fusion rules, end-to-end optimization-driven CNN-based methods were subsequently introduced~\cite{xu2020u2fusion}. These methods optimize similarity-based objectives with respect to the source images, thereby enabling effective integration of complementary source information. Recent architectural advances have provided new impetus for image fusion, leading to the emergence of Transformer-based~\cite{tang2022matr, zhao2023cddfuse, yi2024text}, diffusion-based~\cite{yue2023dif, zhao2023ddfm, yi2024diffif}, and lookup-table-based methods~\cite{yi2025lut} that aim to improve fusion quality, model expressiveness, or efficiency. Meanwhile, considerable progress has been achieved in coupling image fusion with its upstream and downstream tasks~\cite{tang2022superfusion, zhao2023metafusion, yang2026customized}, ranging from prerequisite image registration to subsequent high-level vision applications.

\subsection{Image Registration and Fusion Methods}
To mitigate image misregistration arising from intrinsic discrepancies in multi-modal imaging systems, such as variations in acquisition positions, integrated registration-and-fusion methods have been proposed. RFNet~\cite{xu2022rfnet} integrates style transfer with a stage-wise registration and fusion framework to progressively compensate for parallax-induced spatial misalignment and produce geometrically aligned fused images. Also, UMF-CMGR~\cite{wang2022unsupervised} couples GAN-based cross-modal translation with a multi-level registration and fusion mechanism to facilitate reliable correspondence estimation between initially misaligned multi-modal images. Subsequently, C2RF~\cite{tang2025c2rf} leverages commonality mining and contrastive learning to achieve image decomposition and fusion, while also adapting to subtle geometric deformations. Furthermore, MulFS-CAP ~\cite{li2025mulfs} introduces a learnable modality dictionary and exploits a correlation matrix to jointly perform image registration and fusion in an end-to-end unified framework. FusionRegister~\cite{bian2026fusionregister} follows a strategy that first performs image fusion and subsequently refines severely misaligned regions via targeted correspondence correction, leading to enhanced fusion performance. Despite considerable progress in registration-based image fusion, existing methods remain unable to resolve field-of-view inconsistencies. It further leads to bottlenecks in both visual appearance and semantic information during fusion, thereby limiting the practical applicability of existing methods.

\begin{figure*}[!ht]
\centering
\includegraphics[width=1\linewidth]{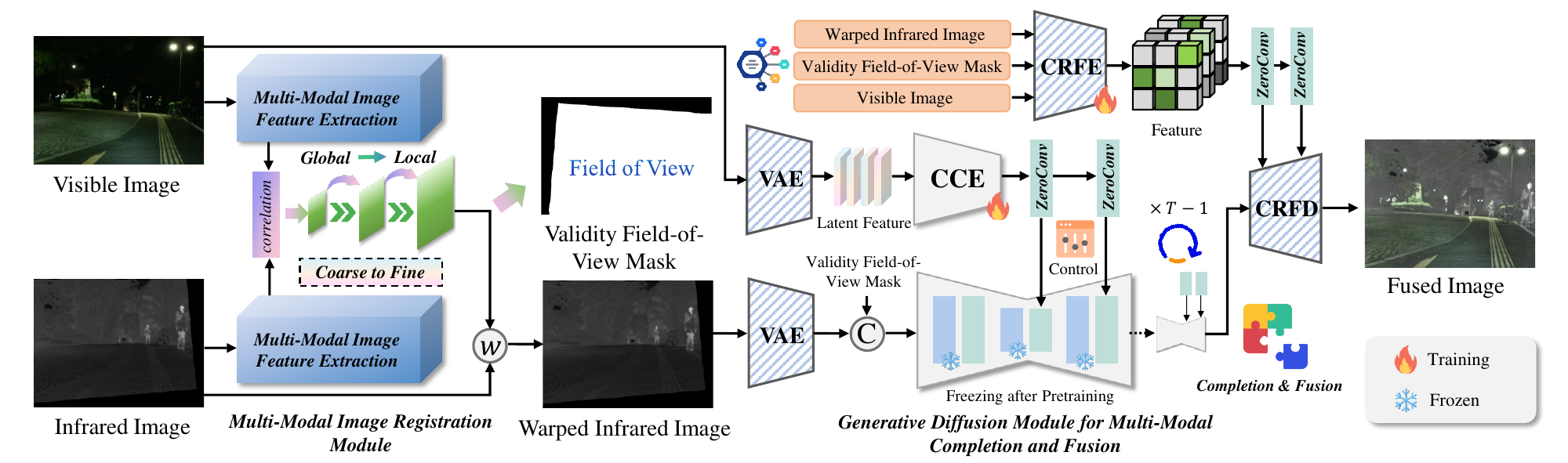}
\caption{The framework of SeeRF. It effectively couples the multi-modal image registration module with the generative diffusion module for multi-modal completion and fusion, leveraging generative modeling to enable cross-modal information fusion beyond conventional shared field-of-view constraints.}
\label{fig3-1}
\end{figure*}

\section{Methodology}
In this section, we first clarify the problem formulation. As illustrated in Fig.~\ref{fig3-1}, we next detail the proposed framework, which comprises a multi-modal image registration module and a generative diffusion module for multi-modal completion and fusion, along with diffeomorphic deformation generation strategy. Finally, we specify the loss functions used to optimize SeeRF.

\subsection{Problem Formulation}
Due to variations in sensor placement and imaging characteristics across multi-modal imaging devices, the acquired source images, \textit{e.g.}, $I_{vis}, I_{ir}$, often suffer from geometric misalignment and field-of-view inconsistencies. Classical image fusion methods typically align the misregistered images into a common geometric space, followed by image fusion. It can be regarded as a function learning problem $\mathcal{F}_{RF}$, and formulated as follows:
\begin{equation}
\label{eq1}
I^{f}=\mathcal{F}_{RF}(I_{vis}, I_{ir}; \theta_{RF}),
\end{equation}
where $\theta_{RF}$ denotes the parameters of image registration and fusion network. $I^{f}$ represents the fusion image. However, this paradigm fails to fundamentally satisfy the full prerequisites of image fusion. Only registration is incapable of recovering information beyond the single-modality field of view, thereby introducing severe artifacts and leading to inconsistent fused representations. To this end, we explore a new paradigm that integrates image registration with generative completion in a collaborative manner, enabling effective utilization of information beyond the field of view. It can be expressed as:
\begin{equation}
\label{eq2}
\widetilde{I}_{ir}, C^{f}=\mathcal{F}_{R}(I_{vis}, I_{ir}; \theta_{R}),
\end{equation}
\begin{equation}
\label{eq3}
I^{f}_{CF}=\mathcal{F}_{CF}(I_{vis}, \widetilde{I}_{ir}, C^{f}; \theta_{CF}),
\end{equation}
where $\mathcal{F}_{R}$ and $\mathcal{F}_{CF}$ denote the image registration and image generative completion fusion mapping function. $\theta_{R}$ and $\theta_{CF}$ are the parameters of the multi-modal image registration module and generative diffusion module for multi-modal completion and fusion. $C^{f}$ is the validity field-of-view mask.

By coupling view registration with field-of-view completion, SeeRF maximizes the utilization of source image information for fusion, leading to improved visual appearance and enhanced performance on downstream tasks.

\subsection{Deformation Generation Strategy}
To simulate sensor misalignment and non-rigid deformations, we construct a global–local deformation model based on diffeomorphic transformations parameterized by a stationary velocity field (SVF). Homography accounts for global projective transformations, while the diffeomorphic component models local deformations, enabling spatially varying non-rigid motion while preserving continuous invertibility. The overall deformation field generation is formulated as follows:
\begin{equation}
\label{eq4}
\varphi=H \circ \phi,
\end{equation}
where $H$ corresponds to the global homography-induced deformation field, and $\phi$ corresponds to the local elastic deformation field parameterized by diffeomorphic transformations. $\circ$ represents the deformation field composition operation.

\noindent
\textbf{Homography Deformation Generation:} Homography transformation models the projection process and produces globally consistent deformations, which can be formulated as follows:
\begin{equation}
\label{eq5}
y\sim H(x),
\end{equation}
where $x\in\Omega\subset\mathbb{R}^{2}$ and $\mathbf{y}\in\Omega_H\subset\mathbb{R}^{2}$ denote the original and transformed pixel coordinates, respectively. $\sim$ indicates equality in the sense of homogeneous coordinates. $H$ represents a randomly sampled homography transformation.

\begin{figure}[!t]
\centering
\includegraphics[width=1\linewidth]{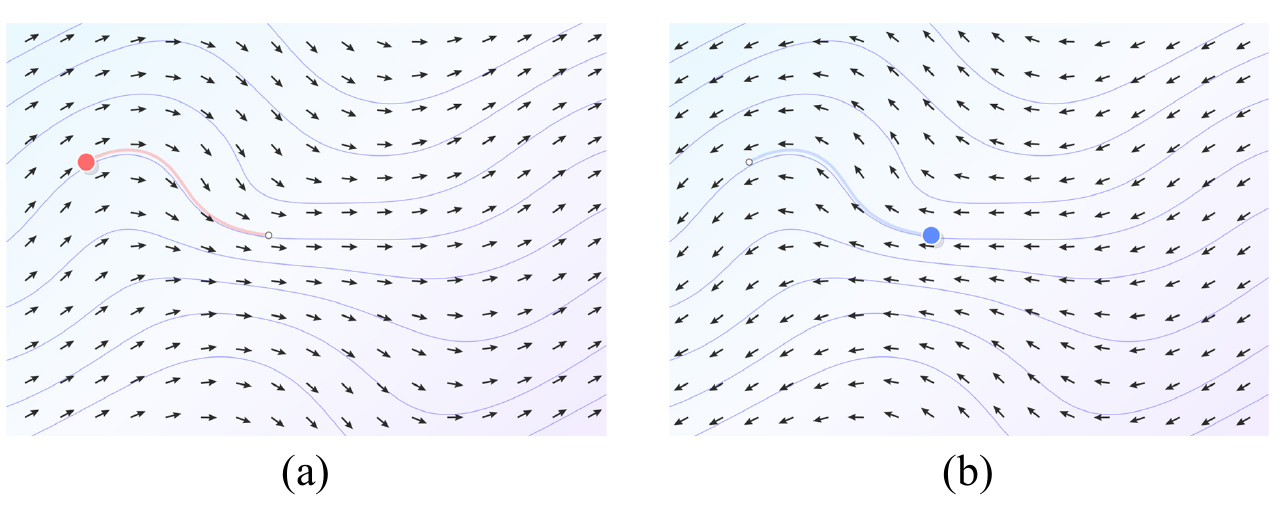}
\caption{Visualization of the stationary velocity field for the diffeomorphic deformation. (a) Motion driven by the forward velocity field; (b) Motion driven by the reverse velocity field.}
\label{fig3-2}
\end{figure}

\noindent
\textbf{Diffeomorphic Deformation Generation:} To address the generated limitations introduced by non-invertible random elastic deformation fields in elastic deformation simulation of mainstream image registration and fusion methods, we propose a diffeomorphic deformation synthesis technique for infrared–visible image registration and fusion, which guarantees the generation of theoretically invertible deformation fields, as shown in Fig.~\ref{fig3-2}. Different from methods that embed diffeomorphic modules into network outputs, we incorporate them into the construction of forward–inverse consistent training samples. Specifically, we formulate elastic deformations for infrared–visible image registration within a stationary velocity field framework. It firstly defines a time-invariant stationary velocity field, defined as follows:
\begin{equation}
\label{eq6}
v(x,t)=v(x), \quad v:\Omega \xrightarrow{} \mathbb{R}^{2}.
\end{equation}

The continuous motion of points in space is governed by a flow equation, from which the relationship between motion and spatial coordinates can be derived as follows:
\begin{equation}
\label{eq7}
\frac{\partial \phi_t(x)}{\partial t} = v\bigl(\phi_t(x)\bigr),
\qquad
\phi_0(x) = x.
\end{equation}

The unit-time deformation flow induced by the stationary velocity field can be defined as the exponential map. Correspondingly, the inverse unit-time deformation flow can be derived by reversing the stationary velocity field. It can be formula as:
\begin{equation}
\label{eq8}
\phi(x)=Exp(v), \qquad \phi(x)^{-1}=Exp(-v),
\end{equation}
where $Exp$ denotes the flow exponential operator rather than an element-wise exponential function. It converts the infinitesimal motion represented by the velocity field into a finite spatial transformation.

Consequently, the forward and inverse deformation fields share the same velocity-field generator. Therefore, they satisfy following equation:
\begin{equation}
\label{eq9}
\phi^{-1}(\phi(x))=x.
\end{equation}

By leveraging the forward and inverse deformation fields, we construct corresponding infrared–visible image pairs with elastic deformations to supervise the learning of registration transformations.

\subsection{Multi-Modal Image Registration Module}
The multi-modal image registration module addresses viewpoint discrepancies in multi-modal source images by integrating vision foundation model guidance with a coarse-to-fine progressive flow regression paradigm, enabling accurate multi-modal registration. Meanwhile, it provides customized registration-confidence masks, which serve as the foundation for completion and fusion, enabling tight coupling between the two processes.

Given unregistered infrared and visible source images, denoted as $I_{ir}$ and $I_{vis}$. Features for registration are extracted from the infrared and visible images using a pretrained pyramid feature extractor, respectively, which can be formulated as follows:
\begin{equation}
\label{eq10}
\{F_{ir,l}^{p}\}^{L}_{l=1}=\mathcal{P}_{i}(I_{ir}), \{F_{vis,l}^{p}\}^{L}_{l=1}=\mathcal{P}_{v}(I_{vis}),
\end{equation}
where $\mathcal{P}_{i}$ and $\mathcal{P}_{v}$ denote the pyramid feature extractors. $\{F_{ir,l}^{p}\}^{L}_{l=1}$ and $\{F_{vis,l}^{p}\}^{L}_{l=1}$ are the extracted features, where $l=1,...,L$ indicates the feature level. Meanwhile, we leverage the vision foundation model (DINOv3) to extract visual features. Benefiting from large-scale pretraining, the model learns highly generalizable representations, which can be formulated as follows:
\begin{equation}
\label{eq11}
\{F_{ir,l}^{n}\}_{l\in S}=\mathcal{N}_{i}(I_{ir}), \{F_{vis,l}^{n}\}_{l\in S}=\mathcal{N}_{v}(I_{vis}),
\end{equation}
where $\mathcal{N}_{i}$ and $\mathcal{N}_{v}$ denote the DINOv3 feature extractors. $\{F_{ir,l}^{n}\}_{l\in S}$ and $\{F_{vis,l}^{n}\}_{l\in S}$ are the extracted foundation features. Considering that the features obtained from vision foundation models are mainly at the semantic level, we combine them with pyramid features at deep level $S$ to obtain more robust representations, expressed as:
\begin{equation}
\label{eq12}
\hat{F}_{x,l}=F_{x,l}^{p}+\mathcal{A}_{l}(F_{x,l}^{p},F_{x,l}^{n}), x\in \{ir,vis\},l\in S,
\end{equation}
where $\mathcal{A}_{l}$ is a gated feature injection mechanism, and $l$ denotes the feature level. It computes attention-based gating weights from the features and modulates feature injection using a temperature parameter:
\begin{equation}
\label{eq13}
\mathcal{A}_{l}\!(F_{x,l}^{p},F_{x,l}^{n})\!=\!\alpha\cdot \sigma\!(\mathcal{L}_{gate}\!(\psi_{c}\!(F_{x,l}^{p},F_{x,l}^{n})))\cdot \mathcal{L}_{proj}\!(\psi_{c}\!(F_{x,l}^{p},F_{x,l}^{n})),
\end{equation}
where $x\in \{ir, vis\}$ and $\alpha$ represent the modality and temperature parameter. $\sigma$ is the Sigmoid function. $\psi_{c}$ denotes the concat operation. $\mathcal{L}_{gate}$ and $\mathcal{L}_{proj}$ denote the gating convolution module and the projection transformation module, respectively, which are used to produce gating coefficients and align feature channels. For the remaining levels, we adopt a feature inheritance strategy from the pyramid:
\begin{equation}
\label{eq14}
\hat{F}_{x,l}=F_{x,l}^{p}, x\in \{ir,vis\},l\notin S.
\end{equation}
Following the paradigm of classical dense matching methods, we adopt a coarse-to-fine flow estimation strategy, which is decomposed into global matching and local matching stages. First, global correlation is computed on deep features to estimate the coarse flow field:
\begin{equation}
\label{eq15}
\mathbf{u}_{L}=\mathcal{M}_{g}(Corr_{global}(\hat{F}_{ir,L},\hat{F}_{vis,L})),
\end{equation}
where $Corr_{global}$ denotes the global correspondence. $\mathcal{M}_{g}$ represents the mapping operation of consistency matching and correlation-to-flow. Subsequently, the coarse flow field is progressively upsampled and refined via local correlation, which can be formulated as:
\begin{equation}
\label{eq16}
\mathbf{u}_{l}=Up(\mathbf{u}_{l+1})+\Delta\mathbf{u}_{l},l=L-1,\dots, 1,
\end{equation}
\begin{equation}
\label{eq17}
\Delta\mathbf{u}_{l}=\mathcal{M}_{l}(Corr_{local}(F_{vis,l},warp(F_{ir,l},Up(\mathbf{u}_{l+1})))),
\end{equation}
where $Corr_{local}$ denotes the local correspondence. $\mathcal{M}_{l}$ represents the mapping operation of local correlation-to-flow. Through progressive iterative refinement, the final flow field is estimated. Therefore, we utilize the full-resolution flow field $\mathbf{u}^{f}$ to warp the source image and obtain the registered image:
\begin{equation}
\label{eq18}
\widetilde{I}_{ir}=warp(I_{ir},\mathbf{u}^{f}).
\end{equation}

To explicitly model the effective overlapping field of view after registration, we estimate a validity field-of-view mask based on the flow field. Specifically, pixel coordinates $\mathbf{p}=(x,y)$ are mapped to new locations using the full-resolution flow field $\mathbf{u}^{f}$:
\begin{equation}
\label{eq19}
q(\mathbf{p})=\mathbf{p}+\mathbf{u}^{f}(\mathbf{p}).
\end{equation}
Subsequently, the validity field-of-view mask is obtained by determining whether the mapped coordinates fall within the valid boundary of the corresponding image:
\begin{equation}
\label{eq20}
C^{f}(\mathbf{p})=
\begin{cases}
1, & 0 \le q_x(\mathbf{p}) < W,\quad 0 \le q_y(\mathbf{p}) < H, \\
0, & \text{otherwise},
\end{cases}
\end{equation}
where $H$ and $W$ are the height and width of the full-resolution image, respectively. In this way, $C^{f}$ can explicitly characterize the geometrically valid region defined by the estimated flow field.

\subsection{Generative Diffusion Module for Multi-Modal Completion and Fusion}
The generative diffusion module aims to address field-of-view inconsistency, which arises from factors such as sensor characteristics and relative sensor configurations. It is tightly coupled with MIR through the registered source image and the validity field-of-view mask, and achieves field-of-view unification and fusion using a diffusion model with both generative and fusion capabilities.

\begin{figure}[!t]
\centering
\includegraphics[width=1\linewidth]{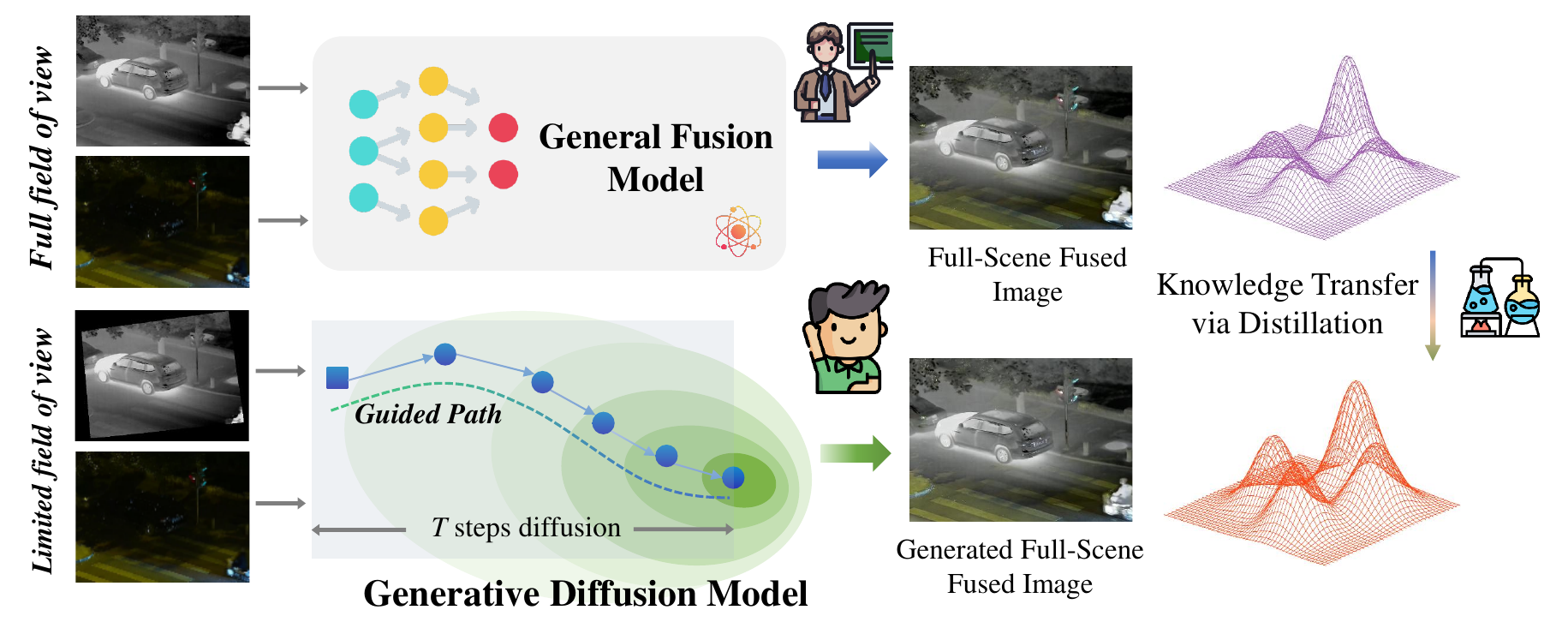}
\caption{Schematic diagram of the generative diffusion module for multi-modal completion and fusion.}
\label{fig3-3}
\end{figure}

We adopt a customized multi-modal latent diffusion model based on the Markov chain formulation of diffusion models~\cite{rombach2022high}. By designing both the forward and reverse diffusion processes, image fusion is achieved. First, the pre-aligned source images with a complete field of view $I_{ir}^{full},I_{vis}^{full}$ is passed through an existing pretrained fusion network $\theta_{f}^{teacher}$ to obtain a full-field fused image:
\begin{equation}
\label{eq21}
I_{f}^{full}=\theta_{f}^{teacher}(I_{ir}^{full},I_{vis}^{full}),
\end{equation}
where $I_{f}^{full}$ is the full-field fused image. The proposed method facilitates prior knowledge transfer through distillation and integration, enabling the generation of complete fusion images covering the entire field of view, as shown in Fig.~\ref{fig3-3}. In detail, it is encoded using a VAE $z_{0}=\mathcal{E}_{full}(I_{f}^{full})$, and the forward diffusion process is formulated as:
\begin{equation}
\label{eq22}
q(z_{t}|z_{0})=\mathcal{N}(z_{t}; \sqrt{\overline{\alpha}_{t}}z_{0}, (1-\overline{\alpha}_{t})\mathbf{I}),
\end{equation}
where $\overline{\alpha}_{t}=\prod_{i=0}^{t} \alpha_{i}, \alpha_{t} = 1-\beta_{t}$, which follows the scheduler of simple diffusion model. $\mathbf{I}$ denotes the identity covariance matrix. Given the registered source image $I_{vis}, \widetilde{I}_{ir}$ and the valid field-of-view mask $C^{f}$, The source images are encoded into a latent space using the encoder of VAEs:
\begin{equation}
\label{eq23}
z_{ir}=\mathcal{E}_{ir}(\widetilde{I}_{ir}),\quad z_{vis}=\mathcal{E}_{vis}(I_{vis}),
\end{equation}
where $z_{ir}$ and $z_{vis}$ denote the latent features in the compressed latent space. $\mathcal{E}_{ir}$ and $\mathcal{E}_{vis}$ are the VAE encoders of the latent diffusion model. Since the registered infrared images typically exhibit a smaller field of view, we treat the infrared image to be completed as the main branch of the diffusion model and fuse its feature information by concatenating it with the valid field-of-view mask. In contrast, the visible image provides consistent spatial structural guidance and is therefore introduced as a control branch through a cross-modal control encoding module into the diffusion process, enabling more accurate cross-modal constraints. The conditional control can be expressed as follows:
\begin{equation}
\label{eq24}
z_{c}=\psi_{c}(z_{ir},C^{f}),
\end{equation}
where $\psi_{c}$ denotes the concat operation. The visible image features are first processed by the cross-modal control encoder to extract multi-level representations $c_{vis}^{l}$, which are then injected into the diffusion denoising network through zero convolutions, enabling visible-guided information generation and fusion:
\begin{equation}
\label{eq25}
\widetilde{h}_{diff}^{l}=h_{diff}^{l}+ZeroConv(c_{vis}^{l}),
\end{equation}
where $h_{diff}^{l}$ denotes the features of the diffusion denoising network at the level $l$. Therefore, the optimization objective of the reverse diffusion process can be expressed as:
\begin{equation}
\label{eq26}
\min \|\epsilon-\epsilon_{\theta}(\sqrt{\overline{\alpha}_{t}} z_{0}+\sqrt{1-\overline{\alpha}_{t}}\epsilon, z_{c}, z_{vis}, t)\|,
\end{equation}
where $\epsilon$ is the noise added by the diffusion model. $\epsilon_{\theta}$ denotes the denoising network. $t$ represents the timestep. The final latent diffusion features are obtained through progressive iterative refinement $\hat{z}_{0}$. To enable the model to learn both unimodal image completion and cross-modal control-guided completion, we first pretrain the diffusion model using full-field-of-view infrared images as the diffusion targets. Subsequently, we freeze the diffusion UNet backbone and train the cross-modal control encoder to generate the full-field-of-view fused images.

To enhance fine-grained texture details in the output fusion image, inspired by ControlNet~\cite{zhang2023adding}, we design a control residual feature decoder (CRFD). Specifically, we duplicate the VAE encoder and make it trainable, termed the control residual feature encoder (CRFE), and inject the extracted features into a frozen pretrained decoder via residual connections and zero convolution. It can be expressed as:
\begin{equation}
\label{eq27}
f_{VAE}^{CRFE}=\mathcal{E}_{CRFE}(\psi_{c}(\widetilde{I}_{ir},I_{vis},C^{f})),
\end{equation}
\begin{equation}
\label{eq28}
I_{f}=\mathcal{D}_{CRFD}^{*}(\hat{z}_{0},f_{VAE}^{CRFE}),
\end{equation}
where $\mathcal{E}_{CRFE}$ denotes the trainable control residual feature encoder. $\mathcal{D}^{*}_{CRFD}$ is the pretrained frozen control residual feature decoder. In particular, the decoder features and the residual injected features are coupled via zero convolution and addition, which can be expressed as:
\begin{equation}
\label{eq29}
\widetilde{h}_{d}^{l}=h_{d}^{l}+ZeroConv(r_{e}^{l}),
\end{equation}
where $h_{d}^{l}$ and $r_{e}^{l}$ denote the features of the control residual feature decoder and $f_{VAE}^{CRFE}$ at level $l$, respectively. $ZeroConv$ is the zero convolution. During training, an approximate output estimation can be obtained using the results predicted by the denoising network:
\begin{equation}
\label{eq30}
\widetilde{z}_{0,t}=\frac{1}{\sqrt{\overline{\alpha}_{t}}}(z_{t}-\sqrt{1-\overline{\alpha}_{t}}\epsilon_{\theta}(z_{t}, z_{c}, z_{vis}, t)).
\end{equation}
Subsequently, the fused image is obtained by estimating the latent representation of the approximate output $\hat{I}_{f}=\mathcal{D}_{CRFD}^{*}(\widetilde{z}_{0,t},f_{VAE}^{CRFE})$, and the optimization objective is defined as:
\begin{equation}
\label{eq31}
\min \|I_{f}^{full}-\hat{I}_{f}\|.
\end{equation}
By jointly optimizing the latent diffusion model and the control residual encoder, we achieve both image fusion and field-of-view consistent completion.

\subsection{Loss Functions}
To effectively optimize the network, we construct the image registration loss and the image fusion loss. The image registration loss is applied to the multi-modal image registration module and is enforced through flow-based constraints. The image fusion loss is applied to the generative diffusion module for multi-modal completion and fusion, and is supervised via the noise prediction and output consistency of the diffusion model.

\noindent
\textbf{For Image Registration:} The flow supervision loss is applied in the image registration stage. By constraining multi-scale flow predictions with the ground-truth flow, the MIR network is optimized. It can be expressed as:
\begin{equation}
\label{eq32}
L_{reg}=\sum_{l=1}^{N}\|\mathbf{u}^{gt}_{l}-\mathbf{u}_{l}\|,
\end{equation}
where $\mathbf{u}^{gt}_{l}$ represents the ground truth flow field at the $l$-th level. Correspondingly, $\mathbf{u}_{l}$ is the estimated flow field predicted by the network at the $l$-th level.

\noindent
\textbf{For Image Fusion:} The latent-space diffusion loss and the fused image consistency loss are introduced to constrain the image fusion results.

In particular, the latent-space diffusion loss enforces supervision on the denoising network output of the diffusion model, enabling generative completion and fusion:
\begin{equation}
\label{eq33}
L_{fus-diff}=\sum_{l=1}^{N}\|\epsilon-\epsilon_{\theta}(\sqrt{\overline{\alpha}_{t}} z_{0}+\sqrt{1-\overline{\alpha}_{t}}\epsilon, z_{c}, z_{vis}, t)\|.
\end{equation}

The fused image consistency loss is constructed by constraining the distillation values between the generated fused image and the full-view fused image:
\begin{equation}
\label{eq34}
L_{fus-VAE}=\|I_{f}^{full}-\hat{I}_{f}\|,
\end{equation}
where $I_{f}^{full}$ denotes the full-view fused image, while $\hat{I}_{f}$ represents the fused image estimated by the network as the completion result.

\section{Experiments}
In this section, we first present the implementation details and the corresponding task settings. We then conduct comprehensive qualitative and quantitative evaluations to validate the effectiveness and superiority of the proposed method. Finally, we perform ablation studies to demonstrate the contribution of each individual module.

\noindent
\textbf{Implementation Details:} We implement the SeeRF with the PyTorch~\cite{paszke2019pytorch} in the machine with four NVIDIA GeForce RTX 4090 GPUs. The optimizer is AdamW~\cite{loshchilov2017decoupled} with a learning rate of 0.0001. For the image registration task, the batch size is 8 and the number of epochs is 800. For the image fusion task, gradient accumulation is used with an effective batch size of 16, and the number of epochs is 400. For the image registration task, we use a patch size of 520 $\times$ 520, while for the image fusion task, the original image resolution is used.

\noindent
\textbf{Datasets:} To validate the effectiveness of the proposed method, we conduct extensive experiments on widely used infrared and visible image fusion datasets, including MFNet~\cite{ha2017mfnet}, FMB~\cite{liu2023multi}, and LLVIP~\cite{jia2021llvip} datasets. For the MFNet dataset, we utilize 784 infrared–visible image pairs for training and 393 pairs for testing and evaluation. For the FMB dataset, 1220 image pairs are used for training, while 280 pairs are reserved for testing. Similarly, for the LLVIP dataset, we employ 1973 image pairs for training and 432 pairs for testing. For deformation generation, we synthesize geometric transformations using homographic transformations and elastic deformations. Specifically, the rotation angle of the homographic transformation is sampled from $[-4^{\circ},4^{\circ}]$, while the horizontal and vertical translations are constrained to 8\% of the image width and height, respectively. For an image of size of $640\times480$, these constraints correspond to maximum displacements of approximately 51 pixels horizontally and 38 pixels vertically. For elastic deformation, the displacement components along both the horizontal and vertical directions are bounded by 5 pixels. Additionally, we conduct experiments on real-world datasets to demonstrate the practicality of the proposed method. Through experiments on relatively comprehensive datasets, the performance of the method can be effectively verified.

\noindent
\textbf{Metric:} We adopt a comprehensive set of evaluation metrics to thoroughly assess the performance of both registration and fusion. For the cross-modal image registration, we employ the end-point error (EPE)~\cite{barron1994performance} metric to evaluate the accuracy of the estimated flow correspondence, which directly measures the displacement error between the predicted and ground-truth flow fields. Additionally, mean absolute error (MAE), mutual information (MI)~\cite{qu2002information}, correlation coefficient (CC)~\cite{pearson1895note}, and structural similarity index (SSIM)~\cite{wang2004image} are used to evaluate the quality of the registered images by comparing them with the ground-truth images. Among these metrics, lower EPE and MAE values indicate better registration accuracy, whereas higher MI, CC, and SSIM values correspond to more precise image alignment. For the image fusion, we employ standard deviation (SD), visual information fidelity (VIF)~\cite{sheikh2005information}, correlation coefficient (CC), edge information preservation metric $Q^{AB/F}$~\cite{xydeas2000objective}, and structural similarity index measure (SSIM) metrics to evaluate the quality of the fused images. Note that all metrics are evaluated using the full-field-of-view source images that cover the entire scene. SD, VIF, CC, $Q^{AB/F}$, and SSIM are all metrics where larger values indicate better fused image quality.

\noindent
\textbf{SOTA Competitors:} To validate the effectiveness of the proposed method, we conduct systematic experiments on multiple public datasets and compare it with existing state-of-the-art approaches. For the image registration task, we select several representative methods as baselines, including RIFT~\cite{li2019rift}, GLUNet~\cite{truong2020glu}, and CrossRAFT~\cite{zhou2022promoting}. For the image fusion task, we compare our method against U2Fusion~\cite{xu2020u2fusion} and DDFM~\cite{zhao2023ddfm}. Furthermore, to comprehensively evaluate the performance of the proposed framework in joint image registration and fusion, we include a broader range of representative and advanced methods for comparison, including  UMF-CMGR~\cite{wang2022unsupervised}, MURF~\cite{xu2023murf}, SuperFusion~\cite{tang2022superfusion}, AUNet~\cite{lu2025net}, MulFS-CAP~\cite{li2025mulfs}, and CPMFusion~\cite{xiao2026cpmfusion}.

\subsection{Results and Analyses}

\begin{figure*}[!ht]
\centering
\includegraphics[width=1\linewidth]{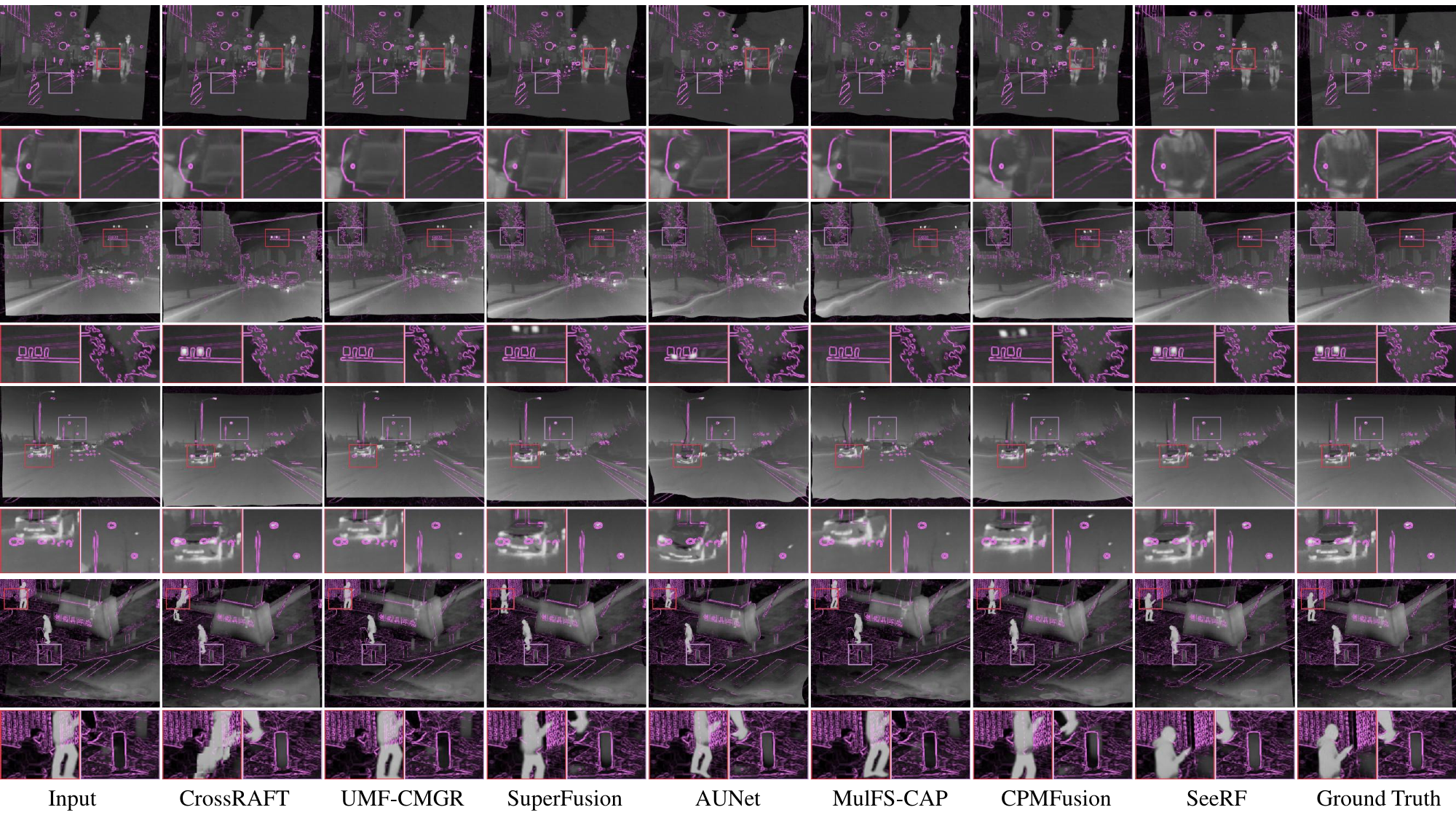}
\caption{Qualitative comparison of SeeRF with state-of-the-art image registration methods on the MFNet (the first group), FMB (the second and third groups), and LLVIP (the fourth and fifth groups) datasets for image registration. We visualize and overlay the gradient of the visible image in purple. Better registration results exhibit higher spatial consistency between the visible image gradients and infrared structures, while maintaining closer object contours and detailed structures compared with the infrared ground-truth images.}
\label{fig4-1}
\end{figure*}

\noindent
\textbf{Qualitative Results on Image Registration:} To provide a more comprehensive evaluation of image registration performance, we visualize the gradient features of the visible images and overlay them onto the registered infrared images for qualitative analysis. The registration quality is assessed from two perspectives: first, the structural similarity between the registered infrared images and the ground-truth infrared images; second, the spatial consistency between salient infrared edges and visible-image gradient responses. A registration method is considered more accurate when the registered results preserve higher structural and detailed consistency with the ground truth while exhibiting stronger spatial correspondence between infrared edge regions and visible gradients, indicating more precise cross-modal spatial alignment. The visualization results are presented in Fig.~\ref{fig4-1}. In the first scene, benefiting from the effective cross-modal registration design, SeeRF accurately aligns the human targets and road edge structures, demonstrating superior registration performance. In contrast, existing competing methods suffer from varying degrees of registration errors, leading to noticeable misalignment in target regions. Such discrepancies are particularly evident in critical structural areas, including human contours and road boundaries, where substantial spatial deviations can be observed. Similarly, in the second and third groups of scenes, CrossRAFT demonstrates relatively strong registration performance in salient regions, including street lamps, vegetation, and vehicles, and is able to establish relatively accurate structural correspondences. However, due to its limited capability in handling complex scenes and boundary regions, its results still exhibit noticeable edge mismatches and local geometric distortions. Constrained by their registration capabilities, UMF-CMGR, MulFS-CAP, and CPMFusion fail to achieve effective alignment in these critical regions, resulting in considerable spatial discrepancies. Although SuperFusion and AUNet can capture certain cross-modal structural consistencies, they still suffer from local misalignment, indicating that their registration accuracy remains suboptimal. In comparison, the proposed method achieves the best performance in both fine-grained object alignment and global spatial consistency. In the fourth scene, CrossRAFT, SuperFusion, and AUNet achieve relatively accurate alignment in the stone pillar regions; however, their spatial correction capability for pedestrian targets remains limited, leading to noticeable deformation and structural distortion in these regions. In contrast, UMF-CMGR, MulFS-CAP, and CPMFusion fail to establish effective cross-modal spatial alignment, resulting in significant registration errors. Benefiting from the proposed registration strategy, SeeRF simultaneously preserves fine-grained object details and global scene structures, achieving superior overall registration performance in these scenarios.

\begin{table*}[!ht]
    \renewcommand{\arraystretch}{1.15}
    \centering
    \caption{Quantitative comparison of our SeeRF with existing state-of-the-art methods of the image registration task on the MFNet, FMB, and LLVIP datasets. (\textbf{Bold}: optimal performance)}
    \resizebox{1\textwidth}{!}{
        \begin{tabular}{c@{\,~~}|c@{\,~~~}c@{\,~~~}c@{\,~~~}c@{\,~~~}c@{\,~~~}
        |c@{\,~~~}c@{\,~~~}c@{\,~~~}c@{\,~~~}c@{\,~~~}|
        c@{\,~~~}c@{\,~~~}c@{\,~~~}c@{\,~~~}c@{\,~~~}}
            \toprule
             \multicolumn{1}{c|}{\multirow{2}{*}{\textbf{Methods}}}& \multicolumn{5}{c|}{\textbf{MFNet Dataset}} & \multicolumn{5}{c|}{\textbf{FMB Dataset}} & \multicolumn{5}{c}{\textbf{LLVIP Dataset}} \\\cline{2-16}
              & EPE & MAE & MI & CC & SSIM & EPE & MAE & MI & CC & SSIM & EPE & MAE & MI & CC & SSIM\\
            \midrule
            RIFT           & 9.9737                                              & 0.0252          & 1.2731          & 0.8014          & 0.8243          & 7.3479          & 0.0246          & 2.1123          & 0.9115          & 0.8354          & 6.3640                      & 0.0456          & 1.3729          & 0.8503          & 0.6646          \\
            GLUNet         & 15.6251                                             & 0.0282          & 1.2056          & 0.7272          & 0.8156          & 11.8658         & 0.0272          & 2.0398          & 0.8933          & 0.8297          & 11.4010                     & 0.0503          & 1.3036          & 0.8292          & 0.6588          \\
            CrossRAFT      & 8.5285                                              & 0.0224          & 1.3560          & 0.7977          & 0.8419          & 10.5699         & 0.0319          & 2.0789          & 0.8523          & 0.8236          & 6.3431                      & 0.0414          & 1.4830          & 0.8736          & 0.7037          \\
            MURF           & 119.0738                                            & 0.1514          & 0.5892          & 0.3273          & 0.5562          & 93.1354         & 0.0958          & 1.2406          & 0.5864          & 0.6507          & 197.5209                    & 0.1419          & 0.5718          & 0.4595          & 0.3734          \\
            UMF-CMGR       & 35.5214                                             & 0.0651          & 0.6568          & 0.2784          & 0.6911          & 44.9324         & 0.0983          & 1.1564          & 0.4800          & 0.6399          & 36.4979                     & 0.1283          & 0.4790          & 0.4109          & 0.3939          \\
            SuperFusion    & 16.4511                                             & 0.0395          & 1.1170          & 0.5449          & 0.7844          & 30.2249         & 0.0685          & 1.4927          & 0.6466          & 0.7091          & 17.2222                     & 0.0797          & 0.9676          & 0.6726          & 0.5499          \\
            AUNet          & 19.7142                                             & 0.0407          & 0.9784          & 0.5424          & 0.7715          & 30.8708         & 0.0666          & 1.4713          & 0.6474          & 0.7172          & 18.6910                     & 0.0841          & 0.8538          & 0.6416          & 0.5244          \\
            MulFS-CAP      & 36.4609 & 0.0643          & 0.6533          & 0.3046          & 0.6903          & 45.7767         & 0.0976          & 1.1451          & 0.4893          & 0.6385          & 37.3990 & 0.1278          & 0.4761          & 0.4159          & 0.3928          \\
            CPMFusion      & 19.0854                                             & 0.0442          & 0.9312          & 0.4944          & 0.7564          & 30.0090         & 0.0704          & 1.4090          & 0.6463          & 0.6964          & 19.0808                     & 0.0910          & 0.7762          & 0.6173          & 0.4906          \\
            \rowcolor{pink!40}
            \textbf{SeeRF} & \textbf{3.8548}                                     & \textbf{0.0119} & \textbf{1.6698} & \textbf{0.9279} & \textbf{0.9010} & \textbf{4.3088} & \textbf{0.0166} & \textbf{2.3674} & \textbf{0.9521} & \textbf{0.8809} & \textbf{2.0404}             & \textbf{0.0221} & \textbf{2.0516} & \textbf{0.9507} & \textbf{0.8524} \\
            \bottomrule
        \end{tabular}
    }\label{quan_reg}
\end{table*}

\noindent
\textbf{Quantitative Results on Image Registration:} The quantitative results are summarized in Tab.~\ref{quan_reg}. The experimental results demonstrate that the proposed SeeRF achieves the best overall performance on the MFNet, FMB, and LLVIP datasets across multiple evaluation metrics, including the optical flow estimation error (EPE) and the MAE, MI, CC, and SSIM metrics between the registration results and reference images. These results indicate that SeeRF is capable of estimating more accurate flow fields and producing registered images with stronger structural consistency with the references. Overall, these experiments validate the superiority of the proposed method for cross-modal image registration and further demonstrate its effectiveness and robustness.

\begin{figure*}[!t]
\centering
\includegraphics[width=1\linewidth]{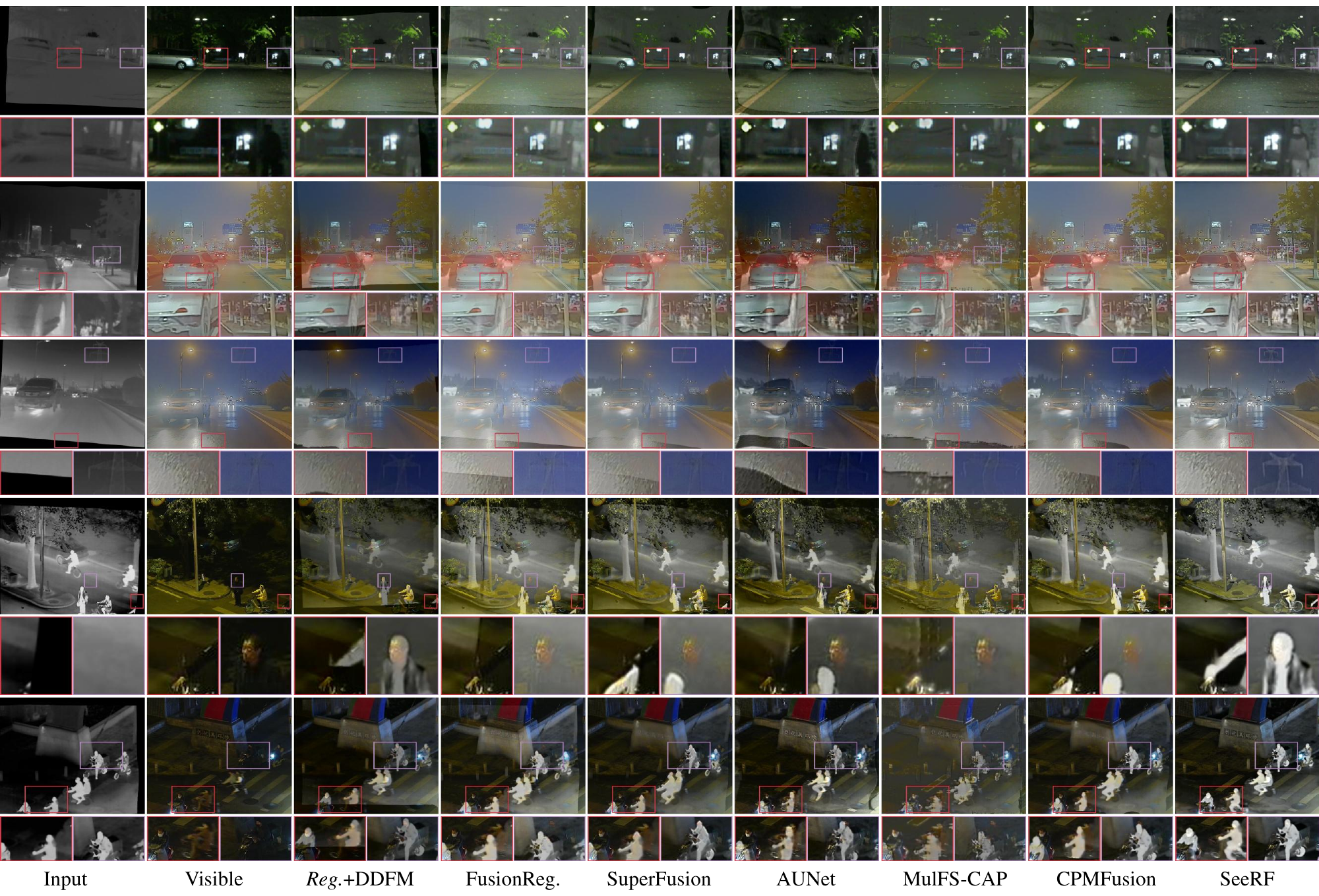}
\caption{Qualitative comparison of SeeRF with state-of-the-art image registration and fusion methods on the MFNet (the first group), FMB (the second and third groups), and LLVIP (the fourth and fifth groups) datasets for image fusion. It is worth noting that the source images are also spatially misaligned.}
\label{fig4-2}
\end{figure*}

\noindent
\textbf{Qualitative Results on Image Fusion:} The corresponding qualitative visualization results are presented in Fig.~\ref{fig4-2}. Our method exhibits two significant advantages. \textit{First, SeeRF is able to generate more complete and natural fused scenes.} Owing to the field-of-view discrepancy between infrared and visible sensors, certain target regions in the registered images may appear in only one modality, which causes existing methods to suffer from information loss, structural discontinuities, and unnatural visual transitions during the fusion process. For instance, in the first scene, part of the infrared information of the pedestrian within the purple box is missing. As most competing methods lack effective generative completion capability, they tend to produce obvious one-sided missing artifacts. Similarly, in non-overlapping field-of-view regions involving structures such as vehicle exhaust pipes, tires, road surfaces, handlebars, and feet, existing methods struggle to adequately model cross-modal field-of-view relationships, often resulting in structural omissions and fusion artifacts. In contrast, by combining an accurate cross-modal registration strategy with a diffusion-based generative fusion mechanism, SeeRF can effectively learn and recover information from non-overlapping field-of-view regions. As a result, it restores more complete target structures while maintaining spatial consistency, thereby producing fused results that are more natural, continuous, and visually coherent. \textit{Second, SeeRF can generate clearer visual representations while effectively preserving key structures and detailed information from the source images.} In the fence region of the first scene and the high-voltage transmission tower region of the third scene, most comparison methods fail to sufficiently preserve texture details and color information from the visible images, while also losing structural edge features of the high-voltage transmission tower, thereby reducing the overall visual quality and scene representation capability of the fused results. In contrast, SeeRF achieves a better balance among texture details, color information, and infrared thermal radiation features during the fusion process, producing high-quality fusion results with clear structures, natural colors, and prominent thermal target information.

\begin{table*}[!t]
    \renewcommand{\arraystretch}{1.15}
    \centering
    \caption{Quantitative comparison of our SeeRF with existing state-of-the-art methods of the image fusion task on the MFNet, FMB, and LLVIP datasets. (\textbf{Bold}: optimal performance)}
    \resizebox{1\textwidth}{!}{
        \begin{tabular}{c@{\,~~}|c@{\,~~~}c@{\,~~~}c@{\,~~~}c@{\,~~~}c@{\,~~~}
        |c@{\,~~~}c@{\,~~~}c@{\,~~~}c@{\,~~~}c@{\,~~~}|
        c@{\,~~~}c@{\,~~~}c@{\,~~~}c@{\,~~~}c@{\,~~~}}
            \toprule
             \multicolumn{1}{c|}{\multirow{2}{*}{\textbf{Methods}}}& \multicolumn{5}{c|}{\textbf{MFNet Dataset}} & \multicolumn{5}{c|}{\textbf{FMB Dataset}} & \multicolumn{5}{c}{\textbf{LLVIP Dataset}} \\\cline{2-16}
              & SD & VIF & CC & $Q^{AB/F}$ & SSIM & SD & VIF & CC & $Q^{AB/F}$ & SSIM & SD & VIF & CC & $Q^{AB/F}$ & SSIM \\
            \midrule
            \textit{Reg}.+U2Fusion & 33.2232          & 0.6445          & 0.5233          & 0.5218          & \textbf{0.7022} & 33.9225          & 0.5727          & 0.5578          & 0.5358          & 0.7107          & 32.2680          & 0.4173          & 0.6164          & 0.3491          & 0.6420          \\
            \textit{Reg}.+DDFM     & 31.5535          & 0.6845          & 0.5237          & 0.5037          & 0.6309          & 30.8141          & 0.4964          & 0.5552          & 0.3663          & 0.6299          & 35.3449          & 0.4631          & 0.6129          & 0.3391          & 0.6489          \\
            FusionReg.     & 34.6501          & 0.5219          & 0.4410          & 0.4728          & 0.5280          & 29.0892          & 0.6338          & 0.5404          & 0.5768          & 0.7611          & 44.9475          & 0.3849          & 0.4727          & 0.3962          & 0.5297          \\
            MURF          & 24.0411          & 0.2911          & 0.3605          & 0.2584          & 0.3905          & 27.7418          & 0.2871          & 0.4605          & 0.3143          & 0.5463          & 26.3277          & 0.1604          & 0.3801          & 0.1586          & 0.3373          \\
            UMF-CMGR      & 24.2902          & 0.4119          & 0.4493          & 0.2914          & 0.4271          & 32.0586          & 0.3703          & 0.4557          & 0.2410          & 0.4628          & 30.9089          & 0.2641          & 0.4449          & 0.2434          & 0.3849          \\
            SuperFusion   & 34.8319          & 0.5855          & 0.5064          & 0.4926          & 0.6079          & 31.6595          & 0.6746          & 0.5466          & 0.5819          & \textbf{0.8306} & 43.4156          & 0.4185          & 0.5452          & 0.3698          & 0.6057          \\
            AUNet         & 27.7622          & 0.5081          & 0.5394          & 0.3423          & 0.4896          & \textbf{36.1326} & 0.3976          & 0.5442          & 0.3635          & 0.4900          & 36.5361          & 0.3592          & 0.5465          & 0.2475          & 0.4751          \\
            MulFS-CAP     & 24.3576          & 0.3118          & 0.4902          & 0.3893          & 0.2992          & 27.9480          & 0.2654          & 0.4967          & 0.3905          & 0.2945          & 31.9229          & 0.2082          & 0.4564          & 0.2941          & 0.3153          \\
            CPMFusion     & 34.7895          & 0.5202          & 0.4926          & 0.4780          & 0.4867          & 34.4127          & 0.6928          & 0.5329          & 0.5716          & 0.7922          & 45.9738          & 0.3772          & 0.5199          & 0.3632          & 0.5029          \\
            \rowcolor{pink!40}
            \textbf{SeeRF (ours)}         & \textbf{35.0617} & \textbf{0.8407} & \textbf{0.5631} & \textbf{0.6372} & 0.7007          & 35.7528          & \textbf{0.8307} & \textbf{0.6312} & \textbf{0.6459} & 0.8086          & \textbf{48.2873} & \textbf{0.7312} & \textbf{0.6366} & \textbf{0.5537} & \textbf{0.8184} \\
            \bottomrule
        \end{tabular}
    }\label{quan_fusion}
\end{table*}

\noindent
\textbf{Quantitative Results on Image Fusion:} The quantitative results are summarized in Tab.~\ref{quan_fusion}. The experimental results demonstrate that SeeRF achieves superior overall performance across multiple image fusion evaluation metrics, validating its effectiveness and robustness under diverse scenes and complex environments. On the MFNet dataset, SeeRF obtains the best performance on four metrics, including SD, VIF, CC, and $Q^{AB/F}$, while maintaining competitive performance in terms of SSIM. Compared with the overall second-best performing method, \textit{Reg.}+U2Fusion, SeeRF achieves improvements of 1.8385, 0.1962, 0.0398, and 0.1154 in SD, VIF, CC, and $Q^{AB/F}$, respectively. These results demonstrate that the proposed method can better preserve salient information from the source images while enhancing the structural consistency and visual quality of the fused results. On the FMB dataset, the proposed SeeRF achieves the best overall performance, obtaining the optimal results on VIF, CC, and $Q^{AB/F}$, while maintaining competitive performance on SD and SSIM. Furthermore, on the LLVIP dataset, SeeRF ranks first across all evaluation metrics, demonstrating stable and consistent performance advantages. These results comprehensively validate the superior capability of SeeRF in enhancing fused-image contrast, preserving visual information fidelity, retaining salient information from source images, and maintaining structural consistency. They further demonstrate the effectiveness and robustness of the proposed method across diverse infrared–visible image fusion scenarios.

\begin{figure*}[!t]
\centering
\includegraphics[width=1\linewidth]{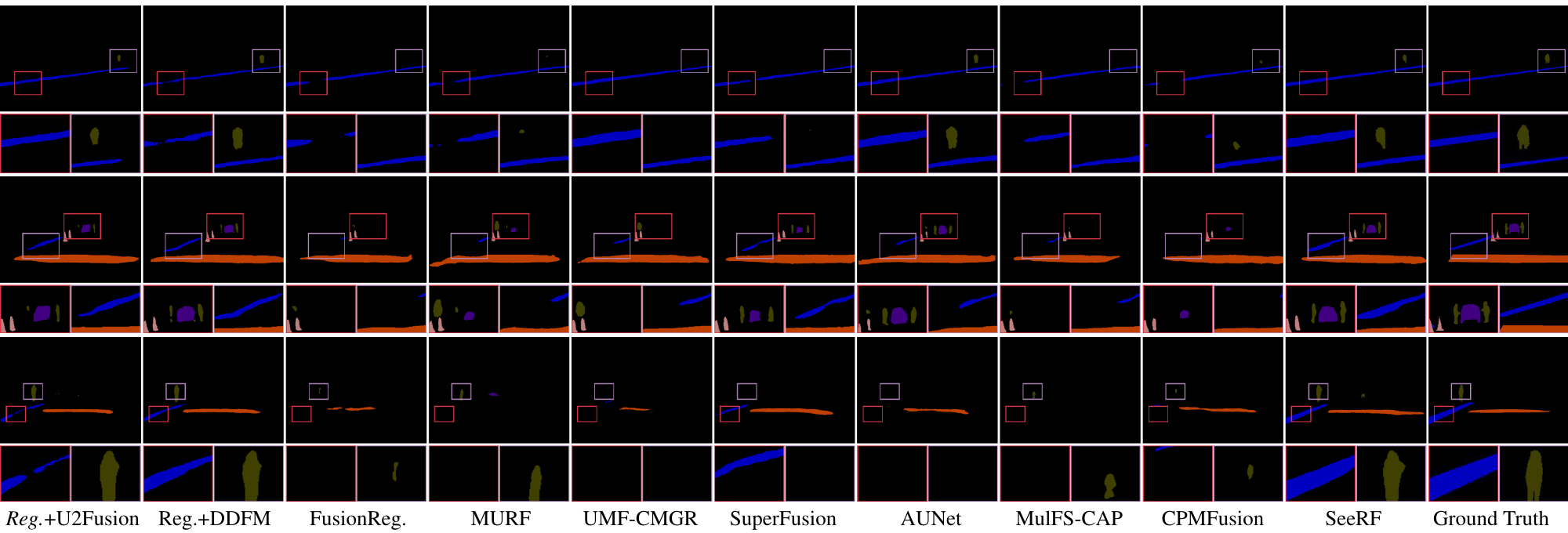}
\caption{Qualitative comparison of SeeRF with state-of-the-art image registration and fusion methods on the MFNet dataset for semantic segmentation task.}
\label{fig4-3-1}
\end{figure*}

\begin{figure*}[!t]
\centering
\includegraphics[width=1\linewidth]{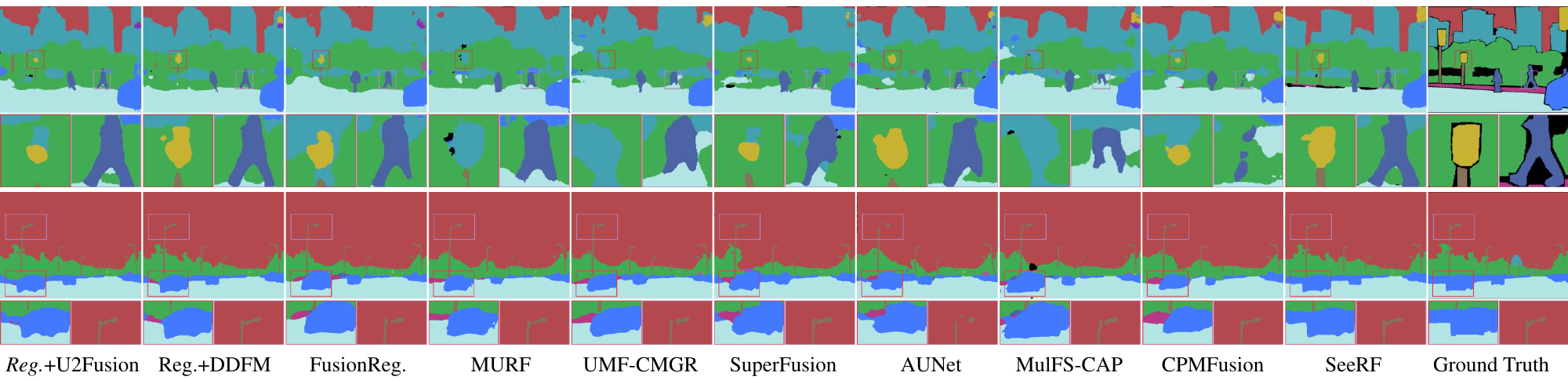}
\caption{Qualitative comparison of SeeRF with state-of-the-art image registration and fusion methods on the FMB dataset for semantic segmentation task.}
\label{fig4-3-2}
\end{figure*}

\begin{table*}[!h]
    \caption{Quantitative results of multi-modal semantic segmentation validation for all categories in MFNet dataset. (\textbf{Bold}: optimal performance)}
    \renewcommand{\arraystretch}{1.15}
    \centering
    \resizebox{0.98\textwidth}{!}{
    \begin{tabularx}{\linewidth}{
        @{}>{\centering\arraybackslash}p{2.0cm}|
        *{9}{>{\centering\arraybackslash}X}|
        >{\centering\arraybackslash}p{1.2cm}@{}
    }
    \toprule
    \textbf{Methods} & Bac. & Car & Per. & Bike & Curve & Car\_stop & Guar. & Col. & Bump & mIoU \\
    \midrule
    \textit{Reg.+U2Fusion} & 97.87 & 86.47 & 63.54 & 62.95 & 31.36 & 28.33 & 4.87 & 52.12 & 50.73 & 53.14          \\
    \textit{Reg.+DDFM}     & 97.91 & 87.84 & 64.83 & 63.20 & 29.92 & 27.92 & 4.27 & 49.67 & 50.85 & 52.93          \\
    FusionReg              & 97.54 & 84.17 & 57.31 & 59.29 & 23.82 & 25.96 & 5.97 & 43.19 & 39.10 & 48.48          \\
    MURF                   & 97.52 & 83.35 & 55.72 & 60.88 & 22.48 & 27.74 & 4.88 & 41.60 & 37.24 & 47.93          \\
    UMF-CMGR               & 97.67 & 84.86 & 58.59 & 61.09 & 27.17 & 27.31 & 3.15 & 42.46 & 47.67 & 50.00          \\
    SuperFusion            & 97.76 & 84.68 & 62.28 & 61.26 & 35.96 & 21.00 & 0.44 & 49.00 & 50.37 & 51.42          \\
    AUNet                  & 97.67 & 84.30 & 60.57 & 61.31 & 28.95 & 27.41 & 7.58 & 49.77 & 47.29 & 51.65          \\
    MulFS-CAP              & 97.52 & 84.04 & 54.76 & 58.23 & 25.29 & 22.20 & 4.95 & 43.77 & 37.85 & 47.62          \\
    CPMFusion              & 97.68 & 84.57 & 59.11 & 59.46 & 29.31 & 16.64 & 2.97 & 45.84 & 41.44 & 48.56          \\
    \rowcolor{pink!40}
    \textbf{SeeRF (ours)}  & 98.08 & 87.43 & 68.85 & 64.00 & 43.24 & 25.03 & 3.17 & 46.16 & 52.50 & \textbf{54.27} \\
    \bottomrule
    \end{tabularx}
    }
    \label{quan_MFNet_image_seg}
\end{table*}

\begin{table*}[!h]
    \caption{Quantitative results of multi-modal semantic segmentation validation for all categories in FMB dataset. (\textbf{Bold}: optimal performance)}
    \renewcommand{\arraystretch}{1.15}
    \centering
    \resizebox{\textwidth}{!}{
    \begin{tabularx}{\linewidth}{
        @{}>{\centering\arraybackslash}p{1.7cm}|
        *{14}{>{\centering\arraybackslash}X}
        >{\centering\arraybackslash}p{0.55cm}|
        >{\centering\arraybackslash}p{0.7cm}@{}
    }
    \toprule
    \textbf{Methods} & Bac. & Road & Sid. & Buil. & Lamp & Sign & Veg. & Sky & Per. & Car & Trunk & Bus & Mot. & Bic. & Pole & mIoU\\
    \midrule
    \textit{Reg}.+U2Fusion & 15.74 & 88.24 & 48.91 & 82.16 & 24.34 & 74.44 & 86.44 & 94.05 & 56.54 & 78.71 & 29.19 & 23.07 & 36.51 & - & 48.51 & 56.20 \\
    \textit{Reg}.+DDFM     & 15.29 & 88.06 & 47.16 & 81.63 & 29.79 & 74.53 & 86.29 & 94.04 & 55.95 & 78.25 & 22.84 & 20.83 & 37.96 & - & 48.89 & 55.82 \\
    FusionReg         & 16.14 & 86.75 & 48.53 & 80.05 & 30.36 & 73.76 & 84.86 & 93.41 & 43.77 & 79.19 & 28.48 & 20.49 & 37.56 & - & 48.06 & 55.10 \\
    MURF              & 16.20 & 87.14 & 45.97 & 78.53 & 22.39 & 72.80 & 84.02 & 92.70 & 44.35 & 77.25 & 18.84 & 14.53 & 32.05 & - & 47.04 & 48.92 \\
    UMF-CMGR          & 15.70 & 87.46 & 46.30 & 78.76 & 26.56 & 72.41 & 84.47 & 93.09 & 43.57 & 78.11 & 21.60 & 16.99 & 37.76 & - & 46.61 & 53.53 \\
    SuperFusion       & 15.77 & 86.53 & 47.39 & 80.81 & 25.31 & 73.55 & 85.05 & 93.47 & 45.20 & 78.32 & 25.82 & 20.44 & 29.04 & - & 48.05 & 53.91 \\
    AUNet             & 16.19 & 86.88 & 46.07 & 79.36 & 28.32 & 72.67 & 84.89 & 93.30 & 44.71 & 78.90 & 24.67 & 13.79 & 34.11 & - & 46.84 & 53.62 \\
    MulFS-CAP         & 14.67 & 86.75 & 42.83 & 76.77 & 28.12 & 72.08 & 83.01 & 92.15 & 38.52 & 77.22 & 21.36 & 13.52 & 26.04 & - & 44.63 & 51.26 \\
    CPMFusion         & 15.46 & 86.40 & 44.17 & 80.45 & 30.25 & 72.75 & 85.32 & 93.68 & 47.98 & 76.10 & 19.90 & 19.20 & 25.96 & - & 47.33 & 53.21 \\
    \rowcolor{pink!40}
    \textbf{SeeRF (ours)} & 16.67 & 88.99 & 48.51 & 82.52 & 33.94 & 74.80 & 86.99 & 94.32 & 61.55 & 79.90 & 30.28 & 21.13 & 35.66 & - & 49.57 & \textbf{57.49} \\
    \bottomrule
    \end{tabularx}
    }
    \label{quan_FMB_image_seg}
\end{table*}

\begin{figure}[!t]
\centering
\includegraphics[width=1\linewidth]{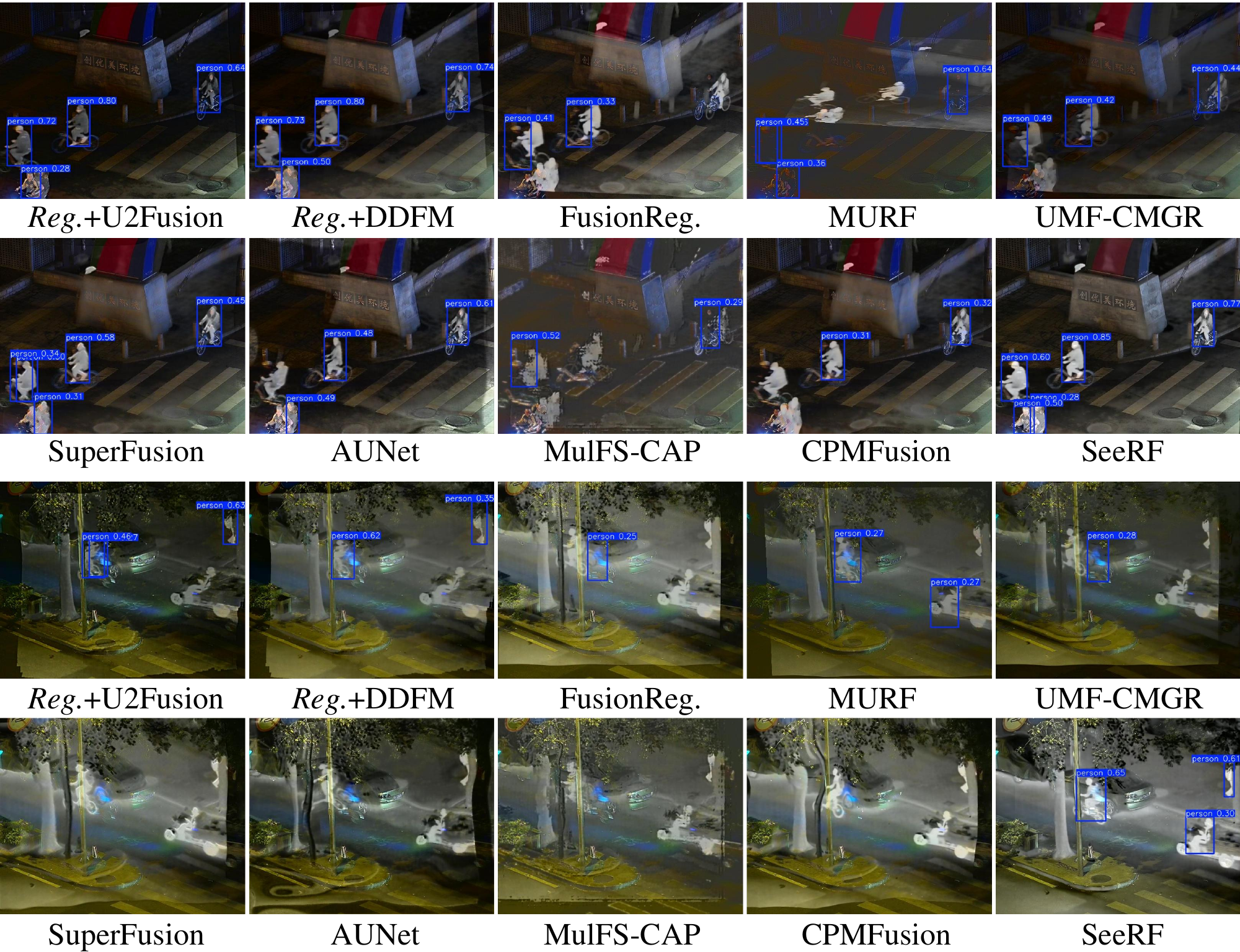}
\caption{Qualitative comparison of SeeRF with state-of-the-art image registration and fusion methods on the LLVIP dataset for object detection task.}
\label{fig4-4}
\end{figure}

\begin{table}[!t]
    \renewcommand{\arraystretch}{1.25}
    \centering
    \caption{Quantitative comparison of object detection on the LLVIP dataset. (\textbf{Bold}: optimal performance)}
    %\vspace{-0.23cm}
    \begin{tabularx}{0.48\textwidth}{
    c
    >{\centering\arraybackslash}X
    >{\centering\arraybackslash}X
    >{\centering\arraybackslash}X
    >{\centering\arraybackslash}X
    }
    \toprule
    \textbf{Methods} & Precision & Recall & mAP@50 & mAP@50:95 \\
    \midrule
    \textit{Reg}.+U2Fusion    & 0.869          & 0.737          & 0.814          & 0.408          \\
    \textit{Reg}.+DDFM        & 0.854          & 0.717          & 0.797          & 0.392          \\
    FusionReg.   & 0.576          & 0.445          & 0.493          & 0.179          \\
    MURF        & 0.688          & 0.532          & 0.596          & 0.246          \\
    UMF-CMGR    & 0.596          & 0.490          & 0.531          & 0.207          \\
    SuperFusion & 0.652          & 0.565          & 0.571          & 0.242          \\
    AUNet       & 0.707          & 0.574          & 0.633          & 0.276          \\
    MulFS-CAP   & 0.527          & 0.368          & 0.395          & 0.137          \\
    CPMFusion   & 0.530          & 0.484          & 0.487          & 0.177          \\
    \rowcolor{pink!40}
    \textbf{SeeRF (ours)}       & \textbf{0.907} & \textbf{0.768} & \textbf{0.871} & \textbf{0.462} \\
    \bottomrule
    \end{tabularx}
    \label{quan_object_detection}
\end{table}

\subsection{Downstream Task Performance Validation}
\noindent
\textbf{Semantic Segmentation:} For the semantic segmentation task, we adopt SegFormer-B2~\cite{xie2021segformer} as the backbone network and conduct experiments on the MFNet and FMB datasets. The qualitative and quantitative experimental results are presented in Figs.~\ref{fig4-3-1}–\ref{fig4-3-2} and Tabs.~\ref{quan_MFNet_image_seg}–\ref{quan_FMB_image_seg}.

From the perspective of qualitative analysis, SeeRF demonstrates superior segmentation accuracy and stronger capability in preserving complete target structures in downstream semantic segmentation tasks. In contrast, due to the limitations of existing methods, competing methods are prone to target region missing, boundary misclassification, and incorrect category predictions, thereby reducing the ability of fused results to represent high-level semantic information.

From the quantitative evaluation results, SeeRF achieves the best mIoU performance on both the FMB and MFNet datasets, demonstrating its effectiveness for downstream semantic segmentation tasks. These results suggest that, benefiting from more accurate cross-modal alignment, effective information integration, and the capability to complete non-overlapping field-of-view regions, SeeRF is able to produce fused representations with richer semantic information and stronger consistency, thereby further enhancing semantic segmentation performance.

\noindent
\textbf{Object Detection:} For the object detection task, we employ YOLOv11\footnote{\url{https://github.com/ultralytics/ultralytics}} as the evaluation backbone and initialize it with pretrained weights. For the fused images produced by each image fusion method, the object detector is retrained on the corresponding datasets to ensure a fair comparison. The qualitative and quantitative results are presented in Fig.~\ref{fig4-4} and Tab.~\ref{quan_object_detection}, respectively. From a qualitative perspective, constrained by limited viewpoint alignment and field-of-view correspondence capabilities, almost all competing methods suffer from missed detections. Moreover, U2Fusion, MURF, and SuperFusion produce duplicate detections and false positives in some sets of results. In contrast, the proposed method achieves the most comprehensive detection performance, demonstrating superior scene representation capability. From a quantitative perspective, the proposed SeeRF consistently achieves the best performance across all evaluation metrics, including Precision, Recall, mAP@50, and mAP@50:95. Compared with the second-best performing method, U2Fusion, SeeRF obtains improvements of 0.038, 0.031, 0.057, and 0.054 in Precision, Recall, mAP@50, and mAP@50:95, respectively. These results clearly demonstrate the effectiveness of the proposed method in facilitating downstream semantic segmentation tasks.

\begin{figure}[!t]
\centering
\includegraphics[width=1\linewidth]{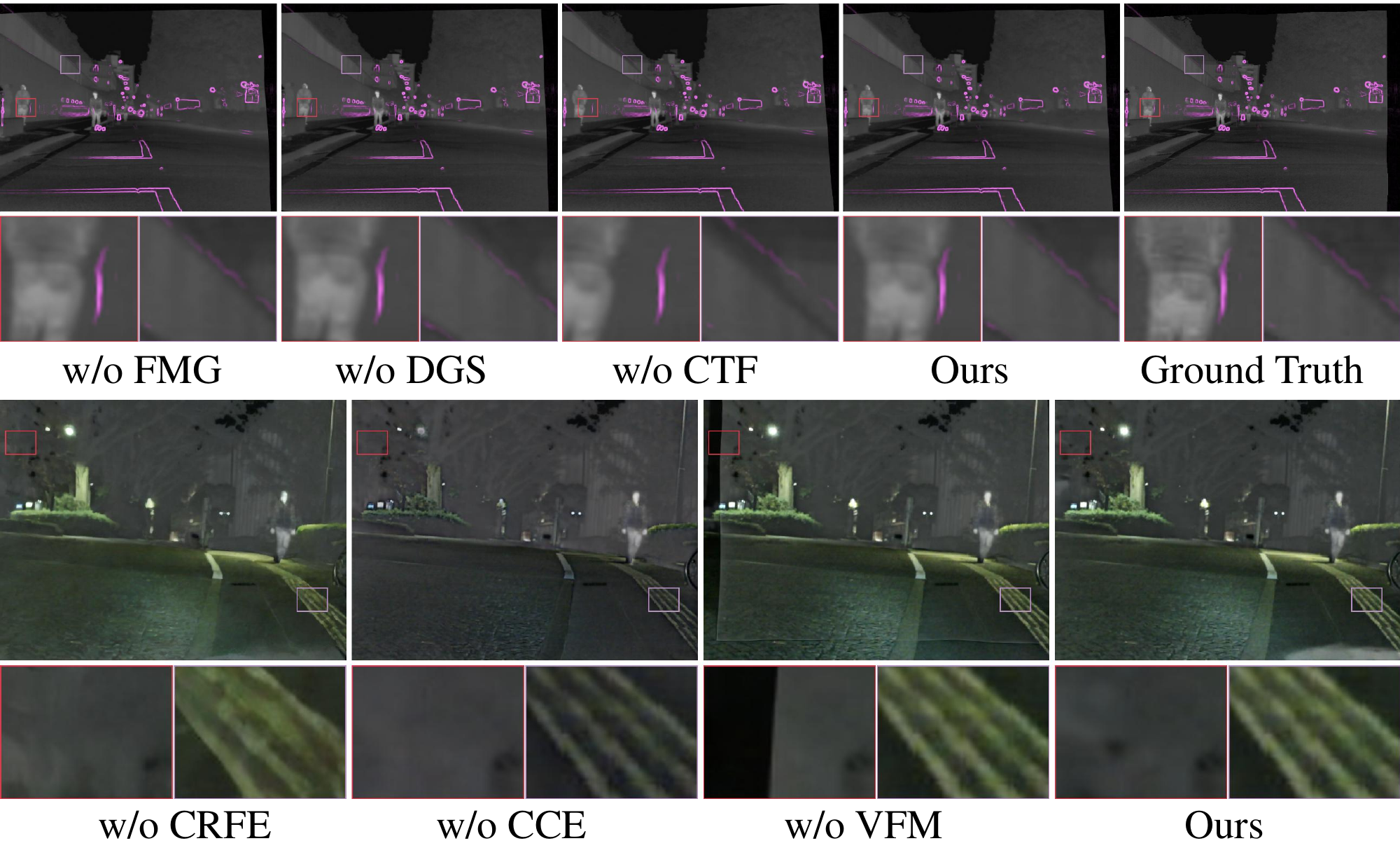}
\caption{Visualization of qualitative comparison results in the ablation study.}
\label{fig4-5}
\end{figure}

\begin{table}[!t]
    \centering
    \renewcommand{\arraystretch}{1.25}
    \caption{Quantitative comparison of the ablation experiment of the general components on MFNet. (\textbf{Bold}: optimal performance)}
    \resizebox{0.49\textwidth}{!}{
    \begin{tabular}{c@{\,~~~~}c@{\,~~~~}c@{\,~~~~}|c@{\,~~~~}c@{\,~~~~}c@{\,~~~~}c@{\,~~~~}c@{\,~~~~}}
        \toprule
        \textit{DGS} & \textit{FMG} & \textit{CTF} & EPE & MAE & MI & CC & SSIM \\
        \midrule
         \checkmark &  & \checkmark & 4.0561 & 0.0127 & 1.6216 & 0.9220 & 0.8944 \\
         & \checkmark & \checkmark & 3.9055 & 0.0132 & 1.6204 & 0.9135 & 0.8910 \\
        \checkmark & \checkmark &  & 10.6553 & 0.0248 & 1.1675 & 0.7887 & 0.8175 \\
        \checkmark & \checkmark & \checkmark & \textbf{3.8548} & \textbf{0.0119} & \textbf{1.6698} & \textbf{0.9279} & \textbf{0.9010} \\

        \midrule
        \midrule
        \textit{VFM} & \textit{CCE} & \textit{CRFE} & SD & VIF & CC & $Q^{AB/F}$ & SSIM  \\
        \midrule
       \checkmark & \checkmark &  & 33.4954 & 0.4360 & 0.5569 & 0.2890 & 0.4169 \\
       \checkmark &  & \checkmark & 21.5992 & 0.5060 & 0.5091 & 0.4544 & 0.5507 \\
        & \checkmark & \checkmark & \textbf{36.1429} & 0.8180 & 0.5460 & 0.6261 & 0.6908 \\
       \checkmark & \checkmark & \checkmark & 35.0617 & \textbf{0.8407} & \textbf{0.5631} & \textbf{0.6372} & \textbf{0.7007} \\

        \bottomrule
    \end{tabular}
    }\label{ablation_quan}
\end{table}

\subsection{Ablation Study}
\noindent
\textbf{General Experiments:} To comprehensively validate the effectiveness of each proposed component, we conduct a series of ablation studies and systematically analyze the contribution of individual components from both image registration and image fusion perspectives. For the MIR, we conduct ablation experiments by removing the foundation model (DINOv3) Guidance mechanism (w/o FMG), replacing the proposed diffeomorphic deformation generation strategy with a conventional random elastic deformation generation method (w/o DGS), and substituting the coarse-to-fine progressive flow regression strategy with direct flow prediction (w/o CTF). For the MCF, we individually remove the control residual feature encoder (w/o CRFE), the cross-modal control encoding module (w/o CCE), and the valid field-of-view mask mechanism (w/o VFM). By ablating the valid field-of-view mask (VFM), we disable the guidance for generative completion, enabling a controlled comparison to evaluate the effect of out-of-field content generation. These experiments aim to systematically investigate the contribution of each component to registration accuracy and fusion quality. The results are presented in Fig.~\ref{fig4-5} and Tab.~\ref{ablation_quan}.

From the perspective of image registration performance, removing any individual component leads to degradation in the final registration results. Specifically, qualitative comparisons show that variants without specific modules tend to exhibit larger spatial offsets, edge misalignment, and structural inconsistencies. Quantitative evaluations further reveal increased in EPE and MAE, accompanied by noticeable declines in MI, CC, and SSIM between the registered and reference images. These results demonstrate the necessity of each proposed component for accurate displacement field modeling, effective cross-modal structural alignment, and improved registration quality.

From the perspective of joint image registration and fusion, qualitative results demonstrate that removing CRFE leads to noticeable structural distortions and loss of fine-grained details in the fused scenes. Without CCE, the weakened cross-modal guidance injection capability results in unnatural color transitions and insufficient information integration. Meanwhile, after removing VFM and preventing the diffusion model from generating content in non-overlapping field-of-view regions, we observe a significant degradation in fusion quality outside the shared field of view, with the overall results exhibiting unnatural discontinuities and incomplete scene representations. Quantitative evaluations further show that, except for the SD metric, the complete model achieves the best performance across all other evaluation metrics. These results further validate the critical contributions of the proposed components in improving fusion quality and enhancing the generative capability of the framework.

\begin{figure}[!t]
\centering
\includegraphics[width=1\linewidth]{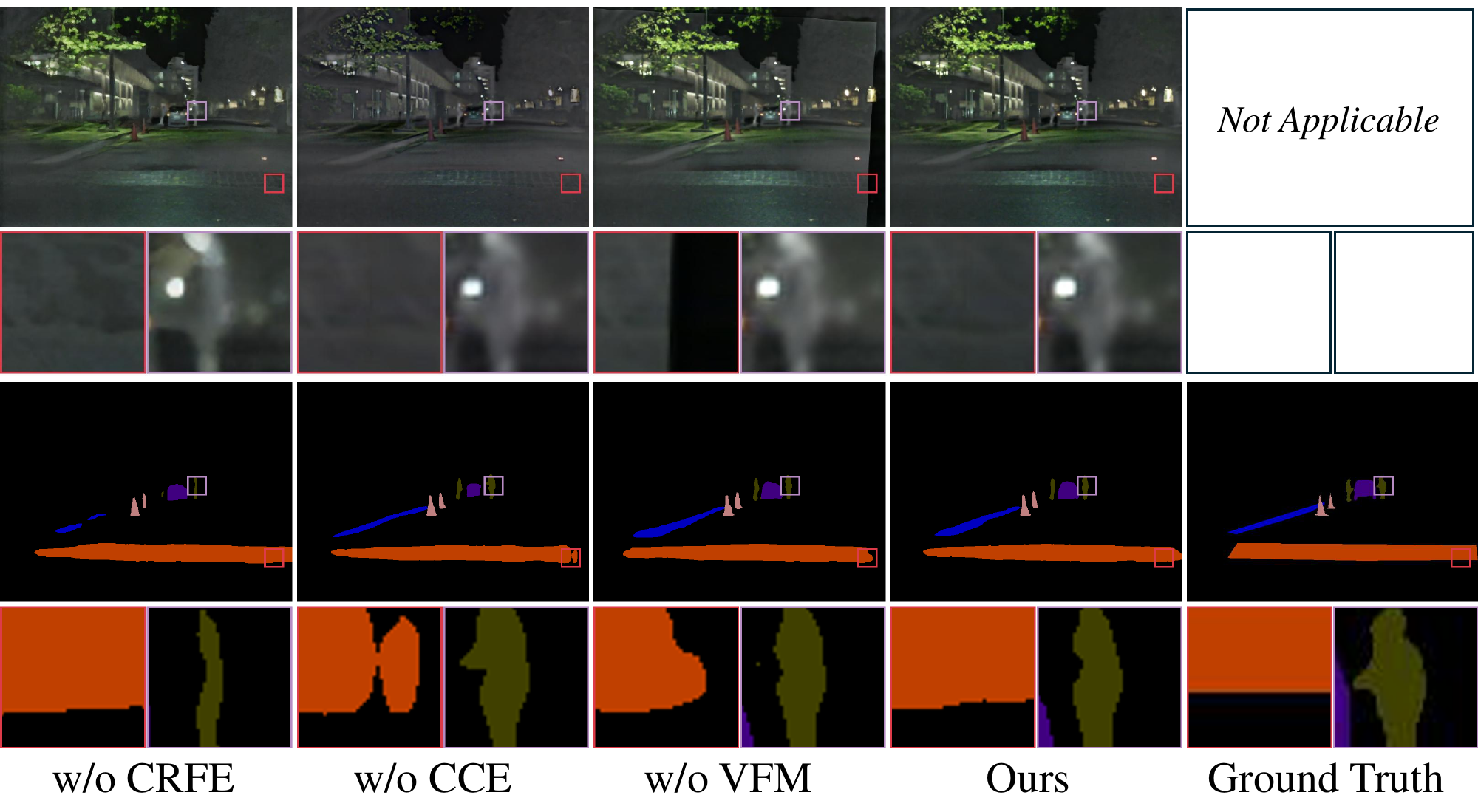}
\caption{Qualitative comparison of semantic segmentation results in the ablation study.}
\label{fig4-6}
\end{figure}

\noindent
\textbf{Semantic Experiments:} To investigate the effects of individual components and out-of-field content generation on downstream semantic segmentation performance, we train all model variants on the MFNet dataset under the same experimental settings as those used in the semantic segmentation task. The corresponding qualitative and quantitative results are presented in Fig.~\ref{fig4-6} and Tab.~\ref{quan_ablation_MFNet_image_seg}, respectively.

From the qualitative results, the absence of modality-specific information in non-overlapping field-of-view regions and the resulting unnatural fusion discontinuities negatively affect the completeness of scene semantic representation. Specifically, the bump target within the red box exhibits obvious structural truncation in the segmentation results, failing to preserve its complete semantic contour. Meanwhile, the corresponding fusion visualization also exhibits poor visual continuity and representation quality. These observations indicate that information loss caused by field-of-view inconsistency not only degrades the visual quality of fused images but also compromises the structural integrity of objects, thereby affecting the accurate recognition and complete representation of targets in downstream semantic segmentation tasks. Similarly, removing either CRFE or CCE impairs the model's ability to capture fine-grained features and effectively integrate multimodal information, consequently degrading its performance on downstream semantic segmentation tasks. These findings further substantiate the motivation behind our approach and demonstrate the important contributions of CRFE and CCE to improving fusion quality and enhancing semantic representation for downstream tasks.

The quantitative results show that disabling content generation beyond the shared field of view leads to a distinct decrease in segmentation mIoU, highlighting the importance of generative completion in addressing field-of-view limitations. Similarly, removing either CRFE or CCE results in degraded semantic segmentation performance, further demonstrating the contributions of these modules to downstream task effectiveness. Overall, the proposed SeeRF achieves the best performance, consistently outperforming all ablation variants by clear margins.

\begin{table*}[!ht]
    \caption{Quantitative results of multi-modal semantic segmentation validation in ablation study for all categories in MFNet dataset. (\textbf{Bold}: optimal performance)}
    \renewcommand{\arraystretch}{1.2}
    \centering
    \resizebox{0.98\textwidth}{!}{
    \begin{tabularx}{\linewidth}{
        @{}>{\centering\arraybackslash}p{2.0cm}|
        *{9}{>{\centering\arraybackslash}X}|
        >{\centering\arraybackslash}p{1.2cm}@{}
    }
    \toprule
    \textbf{Methods} & Bac. & Car & Per. & Bike & Curve & Car\_stop & Guar. & Col. & Bump & mIoU \\
    \midrule
    w/o\_CRFE & 97.81 & 83.91 & 64.41 & 59.86 & 38.87 & 25.73 & 4.00 & 40.16 & 48.90 & 51.52 \\
    w/o\_CCE  & 97.89 & 84.91 & 64.61 & 62.20 & 39.40 & 26.91 & 3.42 & 47.75 & 51.65 & 53.19 \\
    w/o\_VFM  & 97.92 & 85.73 & 67.97 & 62.62 & 38.68 & 24.05 & 3.96 & 47.27 & 54.71 & 53.66 \\
    \rowcolor{pink!40}
    \textbf{SeeRF (ours)}     & 98.08 & 87.43 & 68.85 & 64.00 & 43.24 & 25.03 & 3.17 & 46.16 & 52.50 & \textbf{54.27} \\
    \bottomrule
    \end{tabularx}
    }
    \label{quan_ablation_MFNet_image_seg}
\end{table*}

\subsection{Generative Credibility Analysis Experiment}
To evaluate the feasibility of full-field-of-view image fusion, we independently assess the similarity between the generated content in non-overlapping field-of-view regions and the corresponding ground-truth images from the original modalities. Given the nature of this similarity assessment task, we adopt evaluation metrics similar to those used for image registration and additionally include peak signal-to-noise ratio (PSNR) to quantify the fidelity of the generated content. In the setting where infrared information is missing, poor metric values indicate that the generated regions either lack meaningful infrared information or fail to faithfully represent the actual infrared scene. Conversely, stronger similarity between the generated content and the ground-truth scene corresponds to better metric performance. Accordingly, higher MI, CC, PSNR, and SSIM values indicate more reliable and accurate generation, whereas a lower MAE indicates smaller reconstruction errors and better generation quality.

Building on this, we disable content generation in non-overlapping field-of-view regions (w/o GEN.) by controlling the VFM. Subsequently, we compare the generated content and the non-generative baseline against the corresponding ground-truth regions of the other modality to evaluate the reliability of the generated information. The quantitative experimental results are reported in Tab.~\ref{quan_generation_credibility}. The results demonstrate that the proposed method consistently outperforms the non-generative variant across all quantitative evaluation metrics. This indicates that SeeRF can generate content that is more consistent with the actual scene, thereby effectively improving information representation in non-overlapping field-of-view regions.

\begin{table}[!t]
    \caption{Quantitative results of generative credibility analysis between generated regions and the real modality in the MFNet dataset. (\textbf{Bold}: optimal performance)}
    \renewcommand{\arraystretch}{1.3}
    \centering
    \resizebox{\linewidth}{!}{
    \begin{tabularx}{\linewidth}{
        @{}>{\centering\arraybackslash}p{2.0cm}|
        *{5}{>{\centering\arraybackslash}X}
    }
    \toprule
    \textbf{Methods} & MAE & MI & CC & PSNR & SSIM \\
    \midrule
    w/o GEN. & 0.1379 & 0.5440 & 0.0535 & 15.8329 & 0.3983 \\
    \rowcolor{pink!40}
    \textbf{SeeRF (ours)} & \textbf{0.1067} & \textbf{0.6296} & \textbf{0.2271} & \textbf{18.3355} & \textbf{0.5915} \\
    \bottomrule
    \end{tabularx}
    }
    \label{quan_generation_credibility}
\end{table}

\subsection{Real-World Validation Experiment}
\begin{figure*}[!t]
\centering
\includegraphics[width=1\linewidth]{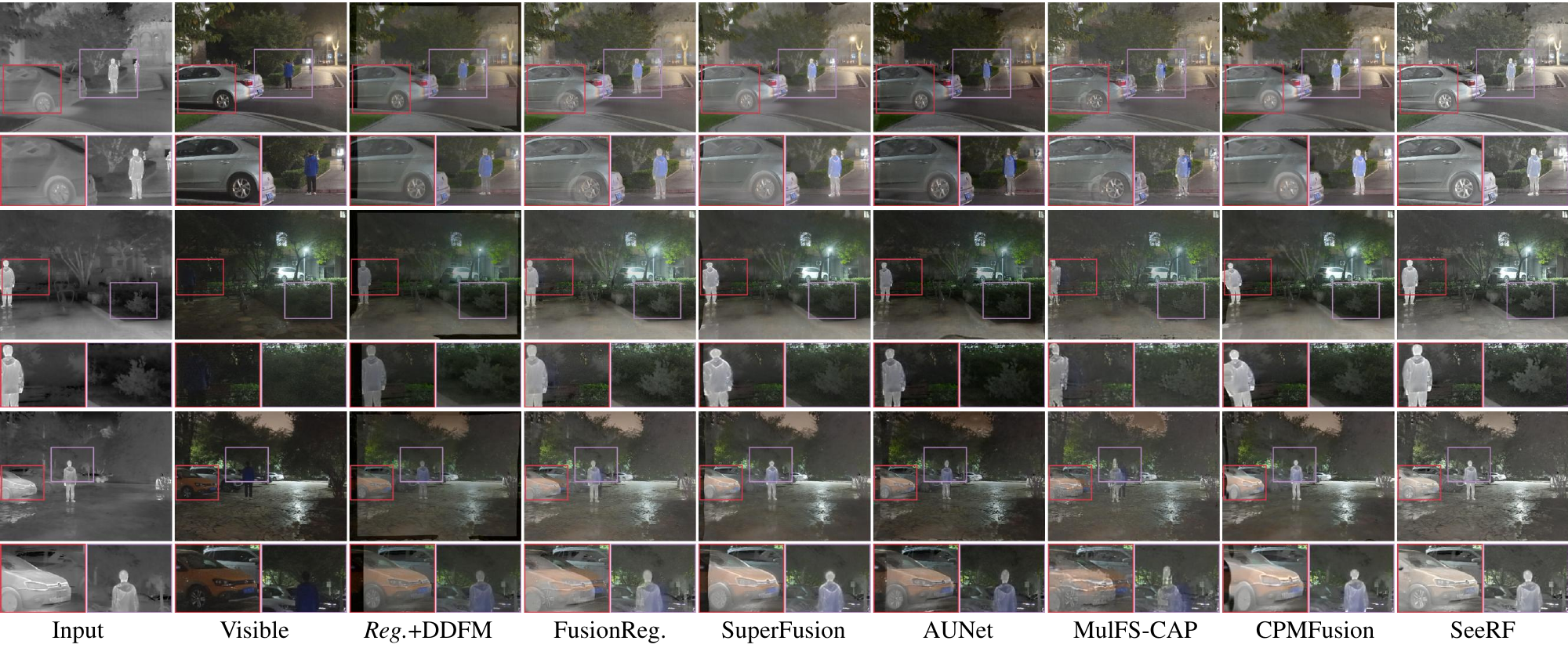}
\caption{Qualitative comparison with the state-of-the-art image registration and fusion methods in the real world.}
\label{fig4-7}
\end{figure*}

To further validate the effectiveness of the proposed method in practical scenarios, we collected real-world data using the multispectral imaging equipment of a DJI Mavic 3T and conducted experiments on the acquired dataset. The corresponding experimental results are presented in Fig.~\ref{fig4-7}.

In real-world scenarios, existing methods exhibit two major limitations. (1) \textbf{Limited registration capability.} Some competing methods struggle to establish accurate cross-modal correspondences, leaving substantial spatial misalignment in the registered results and failing to adequately compensate for target displacements caused by temporal offsets between sensor acquisitions. Specifically, FusionReg and MulFS-CAP fail to achieve effective spatial alignment across all presented scenes, exhibiting pronounced registration failures. CPMFusion, SuperFusion, and AUNet achieve partial improvements in cross-modal alignment; however, noticeable stretching and geometric distortions persist near image boundaries, compromising the overall visual quality of the fused results. Among the competing methods, CrossRAFT, denoted as \textit{Reg.}, demonstrates relatively strong registration performance, achieving reasonable spatial alignment in most scenes, although noticeable misalignment remains in certain boundary regions. By comparison, SeeRF achieves more accurate alignment of both target regions and image boundaries, substantially reducing local misalignment and implausible geometric distortions, thereby delivering superior overall registration performance. (2) \textbf{Field-of-view inconsistency.} All competing methods fail to effectively recover missing information in non-overlapping field-of-view regions. Notably, the limitations imposed by field-of-view discrepancies become even more apparent when accurate spatial alignment has been achieved. Although some methods successfully align the overlapping regions, they fail to adequately recover targets, such as pedestrians and vehicles, located in non-overlapping regions. Consequently, the fused images exhibit pronounced structural discontinuities and unnatural transitions between observed and missing regions. These artifacts compromise the integrity of target-level semantic structures and reduce overall scene fidelity, ultimately limiting the representational capacity of the fused images. In contrast, by leveraging the powerful prior modeling and content completion capabilities of generative diffusion models, SeeRF effectively reconstructs missing content in non-overlapping field-of-view regions and seamlessly integrates the generated content with the observed regions. This enables the generation of more complete and coherent fused images with improved scene representation capability.

\subsection{Computational Overhead}
To evaluate the computational efficiency of the proposed method, we report the number of model parameters, computational complexity (FLOPs), and inference time. Since our fusion process is built upon a diffusion model, we further decompose the entire framework into individual components and analyze their detailed parameter counts and computational costs, enabling a more comprehensive understanding of the resource consumption of each module. The results are reported in Tab.~\ref{computational overhead}. It can be observed that, for the MIR registration component (SeeRF-Reg.), although a visual foundation model is incorporated, the overall computational cost remains acceptable, with 113.367M parameters, 437.667G FLOPs, and an inference time of 0.142s. For the MCF fusion component (SeeRF-Fus.), the overall network is built upon the diffusion model. To preserve strong generative capability, it inevitably introduces higher computational costs, with 1701.765M parameters, 1968.819G FLOPs, and an inference time of 5.293s.

\begin{table}[!t]
    \renewcommand{\arraystretch}{1.4}
    \setlength{\aboverulesep}{0.3ex}
    \setlength{\belowrulesep}{0.3ex}
    \centering
    \caption{Quantitative Computational Overhead Comparison of SeeRF. (Params.: number of parameters, FLOPs: floating-point operations, Time: inference time)}
    \begin{tabularx}{0.5\textwidth}{
    c
    >{\centering\arraybackslash}X
    >{\centering\arraybackslash}X
    >{\centering\arraybackslash}X
    }
    \toprule
    \textbf{Module} & Params (M) & FLOPs (G) & Time (s) \\
    \midrule
    SeeRF-Reg. & 113.367 & 437.667 & 0.142\\
    SeeRF-Fus. & 1701.765 & 1968.819 & 5.293\\
    \bottomrule
    \end{tabularx}
    \label{computational overhead}
\end{table}

\section{Limitation}
To achieve superior registration and fusion performance, we incorporate visual foundation models with strong representation capabilities and diffusion models as the backbone components of our framework. Although these models substantially improve task performance, they inevitably introduce increased computational complexity and inference overhead. Despite employing sampling acceleration strategies to alleviate the computational burden during inference, the proposed architecture still leaves room for further lightweighting and optimization. Future research may explore more efficient generative paradigms, such as single-step generation or flow matching, to reduce computational and memory requirements while preserving the effectiveness of the proposed framework, thereby facilitating its deployment in real-world applications.

\section{Conclusion}
In this paper, we propose a generative registration and fusion framework, termed SeeRF, which aims to overcome the long-standing limitation of shared fields of view in image fusion. It addresses the long-standing issue in image registration and fusion, where existing methods primarily rely on information alignment and fusion within overlapping shared field-of-view regions while neglecting the complementary information contained in sensor-specific regions, resulting in missing information or residual artifacts in non-shared field-of-view regions. Specifically, SeeRF first performs flow field estimation and generates valid field-of-view masks through a diffeomorphic stationary velocity field-based deformation generation strategy, combined with a vision foundation model-guided feature extraction mechanism and a coarse-to-fine progressive multi-modal image registration module. Subsequently, the generative diffusion module for multi-modal completion and fusion is tightly coupled with the multi-modal image registration module, leveraging powerful generative capabilities to overcome the shared field-of-view constraint and achieve effective modeling and fusion of non-shared field-of-view information. Extensive experiments on public datasets and real-world data demonstrate the effectiveness and superiority of the proposed method.

\bibliographystyle{IEEEtran}
\bibliography{egbib}

\end{document}